\documentclass[10pt,twocolumn,letterpaper]{article}

\usepackage{cvpr}              
\definecolor{cvprblue}{rgb}{0.21,0.49,0.74}
\usepackage[pagebackref,breaklinks,colorlinks,allcolors=cvprblue]{hyperref}
\usepackage{booktabs}
\usepackage{multirow}
\usepackage{graphicx}
\usepackage{algorithm}
\usepackage{algpseudocode}

\def\confName{CVPR}
\def\confYear{2027}

\title{Linearized Face Representations for Open-Set Identity Unlearning}

\author{Rouqaiah Al-Refai$^{1}$ \quad Prashant Chandra$^{2}$ \quad Aditya Naresh Patel$^{2}$ \quad Philipp Terh\"{o}rst$^{1}$ \\ 
$^1$ Johannes Gutenberg University Mainz \quad
$^2$ Paderborn University}

\begin{document}
\maketitle
\begin{abstract}
Machine unlearning aims to remove the influence of selected training data without the computational cost of retraining a model from scratch. It has emerged as an important way to address data removal requests and to fulfill data-protection regulations.
This is particularly relevant to face recognition systems, where models are trained on large-scale identity datasets to learn highly discriminative, nonlinear face embeddings, making targeted modifications to identity information difficult.
Previous work on machine unlearning has focused mainly on unrealistic closed-set settings with small-scale datasets and architectures, leaving identity-level unlearning in real-world open-set face recognition underexplored.
In this work, we introduce a linearizer framework that maps face embeddings into a transformed representation space in which a learnable linear transformation enables a direct modification of identity information. Building on this framework, we develop two closed-form unlearning strategies, Identity Singular Subspace Suppression (IS$^3$) and Regularized Identity Suppression (RIS). IS$^3$  identifies the dominant directions that encode a forgotten identity in the linearized representation space and suppresses them, so that the subject's representations collapse toward the subspace shared by all identities. RIS instead redirects forgotten representations toward a predefined target while preserving the original transformation of retained representations. Both strategies are computed in closed form and handle individual as well as simultaneous multi-identity removal requests. We evaluate the linearizer and unlearning pipeline in extensive experiments across multiple face recognition architectures, training objectives, and testing datasets. The results show that linearized representations enable selective and effective identity unlearning in open-set face recognition while largely preserving general recognition performance. Code will be made publicly available.
\end{abstract}    
\section{Introduction}
\label{sec:intro}
Face recognition (FR), a widely deployed biometric technology for authenticating and verifying users' identities, exemplifies the robust, reliable, and sustainable solutions modern security systems require \cite{9481149privacy, DBLP:conf/icb/AlRefaiRRCT25, 10.1145/3736753}.
However, these systems are trained on a large collection of face images, which are privacy-sensitive by nature because of their direct link to a person's identity.
Like other deep neural networks, FR models can memorize and expose personal information without consent \cite{DBLP:conf/icb/TerhorstFDKK20}. Privacy regulations such as Article 17 of the General Data Protection Regulation \cite{eu_gdpr_2016} therefore grant individuals the right to erasure. The most direct way to comply is to retrain the model from scratch without the data to be removed, which is time-consuming and computationally expensive, particularly for large-scale FR systems trained on billions of face samples.
This motivates machine unlearning, which removes the influence of training data without retraining from scratch. 
Current methods \cite{DBLP:journals/tnn/TarunCMK24, DBLP:conf/iclr/FosterSB24, DBLP:conf/aaai/GravesNG21} predominantly aim to approximate the behavior of a model fully retrained without the target data and are mostly developed and evaluated in closed-set settings, such as classification tasks.
In reality, FR is an open-set problem, where the model is trained on a set of known identities, but at deployment it faces unknown identities \cite{DBLP:journals/ivc/VaretoLBSG24}. Consequently, such unlearning methods cannot be transferred directly to settings like open-set FR.
These FR models are trained with deep metric learning objectives that learn face representations either indirectly, through margin-based classification losses whose head is discarded after training (e.g., ArcFace \cite{deng2019arcface}), or directly, by optimizing relative distances between embeddings (e.g., Triplet Loss \cite{Schroff_2015_CVPR}). 
Their open-set nature calls for task-specific unlearning and embedding-level evaluation instead of relying solely on classifier outputs.
The nonlinear, complex design of these architectures therefore motivates closed-form methods that operate on the linear transformation of the learned face representations. Our main contributions are as follows:
\begin{itemize}
    \item \textbf{Linearization framework for open-set FR unlearning.} We introduce a framework that enables efficient identity-level unlearning in pretrained FR models by constructing an editable latent space through an invertible transformation and a learnable linear mapping. To the best of our knowledge, this is the first work to linearize face representations for the purpose of unlearning. This formulation shifts the unlearning operation from modifying large nonlinear backbones to directly editing a compact linear operator, without requiring backbone retraining.
    \item \textbf{Closed-form identity unlearning strategies.} We present two identity unlearning approaches within this framework: Identity Singular Subspace Suppression (IS$^3$), which suppresses dominant forget identity directions in the linear space, and Regularized Identity Suppression (RIS), which modifies the linear mapping by redirecting forget representations toward a common target while preserving the retained ones. Both are formulated for linear identity-level unlearning in open-set FR and modify learned representations directly, without iterative unlearning optimization.
    \item \textbf{Comprehensive evaluation and analysis.} Unlike previous works, which evaluate unlearning on small-scale and closed-set benchmarks, we investigate the proposed framework in realistic large-scale FR settings, covering four state-of-the-art FR training objectives, large-scale backbones, and forgetting requests of varying numbers of identities. Our experiments examine recognition performance preservation, forgetting effectiveness, and scalability, demonstrating effective unlearning for moderate forgetting requests while identifying limitations as the number of forgotten identities increases.
\end{itemize}
\section{Related Work}
\label{sec:related_work}
\paragraph{Machine Unlearning}
Machine unlearning was first introduced by Cao and Yang \cite{7163042towards}, who formed the concept of unlearning as removing the influence of training data without full retraining. 
Ginart \etal later introduced a probabilistic formulation of the concept, which was further developed \cite{DBLP:conf/alt/Neel0S21, DBLP:conf/nips/GuptaJNRSW21, DBLP:conf/nips/SekhariAKS21}.
Machine unlearning methods are broadly categorized as exact and approximate. Exact unlearning provides theoretical guarantees or certificates of removal at the cost of weaker models, whereas approximate unlearning offers a more practical alternative, especially for large scale deep learning models. Most recent methods fall into the latter category, aiming for faster and more practical unlearning while preserving the original model performance.
Proposed unlearning methods target either classification, using strategies such as random labeling, gradient ascent, and Fisher information-based parameter updates \cite{DBLP:conf/cvpr/GolatkarAS20, DBLP:conf/aaai/GravesNG21, DBLP:conf/nips/KurmanjiTHT23, hayase2020selectiveforgettingdeepnetworks}, or generative models, where specific patterns, concepts, or attributes must be removed while preserving overall generative capability \cite{gandikota2024unified, jcp6040121}. A few recent classification oriented methods \cite{DBLP:conf/aaai/FosterSB24, DBLP:conf/iclr/FosterSB24, DBLP:journals/tnn/TarunCMK24} have been evaluated on FR datasets, but mainly in closed-set settings with forget and retain accuracy as metrics, and typically only for small-scale models and single-identity forgetting. Their effectiveness and scalability in open-set FR thus remain largely unexplored.

\paragraph{Machine Unlearning on Face Data}
Machine unlearning has been explored for FR from several perspectives. Most works share the goal of effective unlearning while preserving the model's discriminative ability for identity authentication, but they differ in the task and scenario used for unlearning and evaluation. For instance, several works investigate unlearning in facial analysis tasks beyond FR \cite{DBLP:conf/fgr/GundogduUU24, DBLP:conf/pkdd/HossainMLK25, tok2026nearesttargetwrongone, Choi2023TowardsMU}. Others \cite{DBLP:conf/cikm/LeeW23, 10426386Compromising, DBLP:conf/wacv/Zakharov26, DBLP:conf/icb/ShivamNRG25} study identity unlearning as a closed-set problem, developing methods that assume a final classification head, which is absent in direct representation learning approaches (e.g. triplet loss \cite{Schroff_2015_CVPR}).
Since FR operates in an open-set setting, performance preservation differs from conventional classification. In particular, removing the influence of a previously learned identity should not necessarily prevent the model from verifying pairs of that identity, as a retrained model may still generalize to it as an unseen identity. Accordingly, some approaches, such as \cite{ozturk2026damnatiomemoriaeadversariallyselectively}, intentionally degrade verification performance for forgotten identities, whereas others, such as RAD \cite{alrefai2026unlearningspeedrapidactivation}, aim to preserve the model's verification capability after unlearning. However, RAD requires identifying influential parameters across the network, which introduces computational overhead, and it has only been evaluated for single-identity unlearning. These limitations motivate efficient identity unlearning methods that preserve open-set recognition performance while supporting simultaneous removal of multiple identities.
\section{Methodology}
\label{sec:method}
This paper introduces a framework for linearizing FR models for efficient unlearning by re-expressing their nonlinear embedding transformations through a learnable linear mapping between invertible transformations. This formulation enables direct editing of learned identity features through closed-form operations. Based on that, we propose an Identity Singular Subspace Suppression (IS$^3$) strategy that identifies dominant directions associated with forget identities and projects them away from their identified subspace. We also propose a Regularized Identity Suppression (RIS) strategy that redirects forget identity representations to a target away from the original target subspace while regularizing the mapping to preserve retain representations.

\subsection{Preliminaries and Notation}
\paragraph{Face Recognition.} Consider a deep non-linear FR backbone $M_{\theta}$ parametrized by $\theta$, which maps an input face image $x$ to a $d$-dimensional embedding $e=M_{\theta}(x) \in \mathbb{R}^{d}$. Let $\mathcal{D} = \{(x_{i}, y_{i})\}_{i=1}^{N}$ denote the training dataset, where $y_{i} \in \mathcal{C}$ is the identity label associated with $x_{i}$, and $\mathcal{C}$ is the set of training identities. 
To additionally enable closed-set recognition and unlearning experiments, during training, a classification head $h$ is attached to the backbone to map the learned embeddings $e$ to their identity labels $y$. 
The model is optimized with a supervised objective that encourages intra-class compactness and inter-class separation in the embedding space. After training, the classification head $h$ is retained for closed-set experiments and discarded for open-set ones, where the frozen $M_{\theta}$ serves as the feature extractor.

\paragraph{Identity-Level Unlearning.} Given $\mathcal{C}_f \subseteq \mathcal{C}$ targeted for removal, $\mathcal{D}$ is partitioned into a forget set $D_{f}=\{(x_i,y_i) \in \mathcal{D} \mid y_{i} \in \mathcal{C}_{f} \}$ and a retain set $\mathcal{D}_{r}=\mathcal{D} \setminus \mathcal{D}_{f}$. The objective of identity-level unlearning is to remove the influence of $\mathcal{D}_{f}$ from the learned representations while preserving the original model's recognition capabilities. In other words, the aim is to obtain an updated model $M_{\theta'}$ that behaves similarly to a model $M_{\theta_r}$ retrained from scratch using only $\mathcal{D}_{r}$, such that $M_{\theta'}(x) \approx M_{\theta_r}(x)$ without the expensive computational cost of full-model retraining. This approximation requires accounting for the open-set nature of FR, where identities are verified through embedding similarity rather than pre-defined classes, including independent evaluation datasets $\mathcal{D}_{s}$ unseen during training. Consequently, removing the influence of $\mathcal{D}_{f}$ should not necessarily eliminate the model's ability to verify embedding pairs $e_i'=M_{\theta'}(x_i)$ and $e_j'=M_{\theta'}(x_j)$, belonging to the same forget identity $y_{i}=y_{j} \in \mathcal{C}_{f}$, as a retrained model may still generalize to them as unseen identities. 
\subsection{Post-Hoc Linearizer Framework}
\label{sec:method_lin}
Motivated by \cite{berman2026linearizer}, we propose a post-hoc linearizer framework for FR models to enable direct editing of learned identity information through linear operations. The linearizer framework consists of an invertible transformation $g$ and its inverse $g^{-1}$ wrapped around a learnable linear mapping $A$, and operates on the face embeddings $e=M_{\theta}(x)$, allowing them to be transformed and reconstructed as:
\begin{equation}
    \tilde{e} = g^{-1}(A \, g(e))
\end{equation}
where $\tilde{e}$ is the reconstructed counterpart of $e$.

The invertible transformation $g:\mathbb{R}^{d} \rightarrow \mathbb{R}^{d}$ consists of a stack of $N_b$ coupling blocks, each containing three sequentially applied layers: ActNorm, affine coupling, and permutation. Given $e$, the ActNorm layer applies a learnable feature-wise transformation $u = a \odot e+c$, where $a, c \in \mathbb{R}^{d}$ are learned parameters. The affine coupling layer then splits $u = [u_{1}, u_{2}]$ into two equal parts, where $u_{1}, u_{2} \in \mathbb{R}^{d/2}$. While $u_{1}$ remains unchanged, $u_{2}$ is transformed as $v_{2} = u_{2} \odot \exp(s(u_{1})) + t(u_{1})$, where $s$ and $t$ are learned functions that generate input-dependent scaling and shifting parameters. The resulting $v = [u_{1}, v_{2}]$ is passed through a permutation layer that rearranges its feature dimensions. These operations are repeated across all $N_b$ blocks to obtain $\hat{e} = g(e)$.

These transformed embeddings $\hat{e}$ are passed through a learnable linear mapping $A \in \mathbb{R}^{d \times d}$, producing $A\hat{e}$, which is mapped back to the original embedding space through $g^{-1}$ to obtain the reconstructed embedding $\tilde{e}$. Following \cite{berman2026linearizer}, $A$ is initialized as an identity matrix with small Gaussian perturbations, and the parameters of $g$ and $A$ are jointly optimized to minimize the reconstruction error between $e$ and $\tilde{e}$ using the following objective:
\begin{equation}
    \mathcal{L} = \underbrace{\| \tilde{e} -e\|_{2}}_{\mathcal{L_{\text{rec}}}} + 
    \underbrace{(1 - \cos(\tilde{e},e))}_{\mathcal{L_{\text{cos}}}} + \underbrace{0.1 \| g^{-1}(g(e)) - e \|_{2}}_{\mathcal{L}_{\text{inv}}}
\end{equation}
\noindent which combines an $\ell_{2}$ reconstruction loss $\mathcal{L}_{\text{rec}}$ and a cosine similarity loss $\mathcal{L}_{\text{cos}}$ to preserve the original embedding representations, together with an invertibility regularization term $\mathcal{L}_{\text{inv}}$ that penalizes numerical inversion errors in recovering $e$ through $g^{-1}$. After training, the linearizer parameters are frozen, and only the linear mapping $A$ is modified during unlearning using the proposed closed-form suppression strategies IS$^3$ and RIS. 
\subsection{Identity Singular Subspace Suppression (IS$^3$)}
\label{sec:method_is3}
The first proposed unlearning strategy aims to identify and suppress dominant directions of variation associated with $\mathcal{C}_{f}$ in the transformed linearized embedding space. Given $\mathcal{D}_{f}$, the corresponding face embeddings $e_{i}=M_{\theta}(x_i)$ are first mapped through $g$ to obtain $\hat{e}_{i}=g(e_{i})$, for all $(x_i, y_i) \in \mathcal{D}_{f}$. These embeddings are collected into a matrix $Z_{f} \in \mathbb{R}^{n_f\times d}$, where $n_f = |\mathcal{D}_{f}|$. The matrix is then centered by subtracting the mean transformed embedding $\mu_{f} = \frac{1}{n_{f}}\sum_{i=1}^{n_{f}} \hat{e}_{i}$, yielding $Z_{\mathrm{cen}} = Z_{f}-\mathbf{1}_{n_{f}}\mu_{f}^{\top}$. This process removes the mean component of the forget embeddings, ensuring that the subsequent decomposition captures directions of variation around the mean rather than directions dominated by the mean itself.

Singular Value Decomposition (SVD) is then applied to the centered matrix, obtaining $Z_{\mathrm{cen}} = U\Sigma V^{\top}$, where $U$ and $V$ contain the left and right singular vectors, respectively, and $\Sigma$ contains the singular values in descending order. The first $k$ columns of $V$, denoted by $V_{k} \in\mathbb{R}^{d\times k}$, represent the directions capturing the greatest variation among the centered forget embeddings. Using the selected singular vectors, we construct a rank-$k$ orthogonal projector $P=V_{k}V_{k}^{\top} \in \mathbb{R}^{d\times d}$ into the identified subspace. This projector isolates the components of $\hat{e}$ that lie along the selected directions, which are subsequently targeted for suppression.

To suppress the contribution of the selected directions, the linear mapping $A$ is modified as:
\begin{equation}
    A' = A-\alpha AP
\end{equation}
\noindent where $\alpha \in [0,1]$ controls the suppression strength. Setting $\alpha=1$ completely removes the contribution of the selected subspace, while smaller values partially suppress it. The resulting mapping $A'$ replaces $A$ in the linearizer.
The complete unlearning process is summarized in Algorithm \ref{alg:is3}.

\begin{algorithm}[t]
\caption{Identity Singular Subspace Suppression (IS$^3$)}
\label{alg:is3}
\begin{algorithmic}[1]
\Require $(g,A)$, $\{e_i\}_{i=1}^{n_f}$, strength $\alpha$, rank $k$
\Ensure Edited matrix $A'$
\State $\hat{e}_i \gets g(e_i), \quad \forall i=1,\ldots,n_f$
\State $Z_f \gets [\hat{e}_i^\top]_{i=1}^{n_f}$
\State $\mu_f \gets \operatorname{mean}(Z_f)$
\State $Z_{\mathrm{cen}} \gets Z_f-\mathbf{1}_{n_f}\mu_f^\top$
\State $U,\Sigma,V^\top \gets \operatorname{SVD}(Z_{\mathrm{cen}})$
\State $V_k \gets V_{:,1:k}$
\State $P \gets V_k V_k^\top$
\State $A' \gets A-\alpha AP$
\State \Return $A'$
\end{algorithmic}
\end{algorithm}

\subsection{Regularized Identity Suppression (RIS)}
\label{sec:method_ris}
The second unlearning strategy adapts a similar approach introduced in \cite{gandikota2024unified} to suppress the representation of forget identities while minimizing changes to the remaining representations. Unlike IS$^3$, which suppresses selected directions in the transformed embedding space, RIS modifies $A$ by redirecting forget representations toward a common target while minimizing changes to the retain representations.

Using the transformed face embeddings $\hat{e}_i=g(e_i)$ for $(x_i,y_i)\in\mathcal{D}_f$ and $\hat{e}_j=g(e_j)$ for $(x_j,y_j)\in\mathcal{D}_{r}^{(s)}$, where $\mathcal{D}_r^{(s)}\subseteq\mathcal{D}_{r}$ is a sampled retain set of size $n_r$. The mean transformed retain embedding is computed as $\mu_r=\frac{1}{n_r}\sum_{j=1}^{n_r}\hat{e}_j$, and used to define a common target $z^\star=A\mu_r$. This target represents the original mapping of $\mu_{r}$ through $A$, toward which the forget embeddings are redirected.

To achieve an updated $A'$, RIS aims to minimize the deviation of $\hat{e}_i$ from $z^\star$ while penalizing changes to the original $\hat{e}_j$ mappings. These competing objectives are formulated as a weighted least-squares problem, yielding the following regularized closed-form solution:
\begin{equation}
\begin{aligned}
A' &= BG^{-1},  \qquad \text{with} \\
B &= z^{\star}\mu_f^{\top}
    + \frac{\lambda}{n_r}\sum_{j=1}^{n_r}
    A\hat{e}_j\hat{e}_j^{\top}, \\
G &= \frac{1}{n_f}\sum_{i=1}^{n_f}
    \hat{e}_i\hat{e}_i^{\top}
    + \frac{\lambda}{n_r}\sum_{j=1}^{n_r}
    \hat{e}_j\hat{e}_j^{\top}
    + \varepsilon I.
\end{aligned}
\end{equation}
Here, $\mu_f=\frac{1}{n_f}\sum_{i=1}^{n_f}\hat{e}_i$ denotes the mean transformed forget embedding, while $\lambda\geq0$ controls the trade-off between unlearning and retain preservation. The regularization parameter $\varepsilon>0$ prevents $G$ from becoming singular and is set to $2\times10^{-3}$ times its mean diagonal before regularization to ensure scale-adaptive regularization. Finally, the resulting matrix $A'$ replaces $A$ in the linearizer. The complete unlearning strategy is shown in Algorithm \ref{alg:ris}.

\begin{algorithm}[t]
\caption{Regularized Identity Suppression (RIS)}
\label{alg:ris}
\begin{algorithmic}[1]
\Require $(g,A)$, $\{e_i\}_{i=1}^{n_f}$, $\{e_j\}_{j=1}^{n_r}$, $\lambda$
\Ensure Edited matrix $A'$
\State $\hat{e}_i \gets g(e_i), \quad \forall i=1,\ldots,n_f$
\State $\hat{e}_j \gets g(e_j), \quad \forall j=1,\ldots,n_r$
\State $\mu_f \gets \operatorname{mean}(\hat{e}_i), \quad
       \mu_r \gets \operatorname{mean}(\hat{e}_j)$
\State $z^\star \gets A\mu_r$
\State $B \gets z^\star\mu_f^\top
       + \frac{\lambda}{n_r}\sum_{j=1}^{n_r}
       A\hat{e}_j\hat{e}_j^\top$
\State $G \gets \frac{1}{n_f}\sum_{i=1}^{n_f}
       \hat{e}_i\hat{e}_i^\top
       + \frac{\lambda}{n_r}\sum_{j=1}^{n_r}
       \hat{e}_j\hat{e}_j^\top+\varepsilon I$
\State $A' \gets BG^{-1}$
\State \Return $A'$
\end{algorithmic}
\end{algorithm}

\section{Experimental Setup}
\label{sec:setup}

\subsection{Datasets}
Experiments use the refined MS1M-V2 dataset \cite{MS_celeb_1m}, comprising roughly 5.8 million face images of 85K identities, as the primary training and unlearning dataset. The linearizer framework is trained on a random subset of 5K identities with approximately 20 images each. This subset also provides the identities from which the retain and forget sets, $\mathcal{D}_{r}$ and $\mathcal{D}_{f}$, are constructed, with unlearning requests ranging from a single identity to 1K identities.
Recognition performance before and after unlearning is measured on five widely adopted verification benchmarks: LFW \cite{huang_inria-00321923}, AgeDB-30 \cite{moschoglou2017agedb}, CFP-FP \cite{cfpfp2016Sengupta}, CPLFW \cite{CPLFWTech} and CALFW \cite{abs-1708-08197}. These cover variations in pose, age and appearance under standardized verification protocols.
The evaluation is extended to IJB-B \cite{WhitelamTBMAMKJ17} and IJB-C \cite{MazeADKMO0NACG18}, which exhibit greater variability in image quality, pose and acquisition conditions, using the established 1:1 verification protocols to assess performance under more realistic conditions.

\subsection{Models}
The framework is deployed and the unlearning methods are evaluated on several pretrained FR models, namely ArcFace \cite{deng2019arcface}, AdaFace \cite{9880230}, MagFace \cite{meng2021magface}, QMagFace \cite{10030931} and DQM-Face \cite{DBLP:conf/eccv/BelabbaciWT26}, all trained on MS1M-V2. To assess generalizability across model capacities, two backbone sizes are considered, iResNet50 and iResNet100, except for DQM-Face, for which only iResNet100 is available. Results for iResNet50 are reported in the supplementary, as they show similar trends to iResNet100 without providing additional insights. These are large-scale models with substantially more parameters than the smaller architectures, such as ResNet18, commonly used in machine unlearning studies. Unlike conventional classification models, FR models produce identity discriminative embeddings that enable open-set verification at inference. Although the proposed methods operate directly on these embeddings, the pretrained classification head weights are retained at test time to examine closed-set performance, in line with evaluation practices adopted in existing machine unlearning studies.

\subsection{Evaluation Metrics}
\label{sec:eval_metrics}
\begin{table}[t]
\footnotesize
\centering
\caption{\textbf{Hyperparameter configurations.} The proposed unlearning methods across different numbers of forgotten identities.}
\label{tab:hyperparameters}
\setlength{\tabcolsep}{5pt}
\renewcommand{\arraystretch}{0.8}
\begin{tabular}{llccccccc}
\toprule
& $|\mathcal{C}_f|$ & \textbf{1} & \textbf{5} & \textbf{10}
& \textbf{50} & \textbf{100}
& \textbf{500} & \textbf{1K} \\
\midrule
\multirow{2}{*}{IS$^3$}
& $\alpha$ & 0.75 & 0.99 & 0.99 & 0.98 & 0.98 & 0.99 & 0.99 \\
& $k$      & 2 & 5 & 20 & 50 & 150 & 256 & 256 \\
\midrule
RIS
& $\lambda$ & 5 & 1 & 0.5 & 0.1 & 0.05 & 0.05 & 0.05 \\
\bottomrule
\end{tabular}
\end{table}

Face verification performance is reported as mean accuracy over ten folds for each of the five standard benchmarks, under the protocol of each benchmark. On IJB-B and IJB-C, the True Accept Rate (TAR) is reported at multiple predefined False Accept Rate (FAR) thresholds, following established biometric evaluation protocols \cite{ISO_IEC_19795_1_2006}. Unlearning effectiveness in the open-set setting is measured with the Forget Score (FS) of \cite{alrefai2026unlearningspeedrapidactivation}, which quantifies relative changes in the embeddings of forget and retain samples before and after unlearning and is the only open-set unlearning metric available for FR. For the closed-set setting, forget and retain accuracies are computed with the corresponding pretrained classification heads, aligning with common practice in machine unlearning research despite the limited applicability of closed-set evaluation to open-set face recognition. Finally, the execution time of each method is reported to analyze computational efficiency.

\subsection{Implementation and Hyperparameter Tuning}
\label{sec:implementation}

In the experiments, publicly available implementations and pretrained weights are used for each face recognition model. The proposed linearizer framework for identity-unlearning consists of $N_{b}=6$ blocks and is trained with AdamW for $30$ epochs at a learning rate of $10^{-4}$. To ensure that it preserves the original embedding representations, its reconstruction fidelity and invertibility are verified. A linearizer is accepted when its construction cosine similarity exceeds $0.99$ and its inversion error remains below $10^{-6}$.

Identity level unlearning is evaluated at seven scales, with the number of forgotten identities $|\mathcal{C}_{f}| \in \{1, 5, 10, 50, 100, 500, 1000\}$. For each scale, disjoint identity partitions are constructed and kept fixed across methods to ensure consistent conditions. IS$^3$ is controlled by $\alpha$ and $k$, and RIS by $\lambda$. Both hyperparameters are tuned per unlearning scale and monitored consistently across all FR models. Table~\ref{tab:hyperparameters} reports selected values.

For comparison, the pretrained models are reported as Base and their linearized counterparts as Lin. RAD and LFSSD-E \cite{alrefai2026unlearningspeedrapidactivation} are additionally reproduced with the corresponding iResNet100 pretrained models, following the hyperparameter configurations reported in their work. Unlike \cite{alrefai2026unlearningspeedrapidactivation}, which averages results over five independently forgotten identities, all methods are evaluated on a randomly selected identity in a single unlearning run per model, ensuring a consistent single identity comparison.

\section{Results and Discussion}
\label{sec:results}

\begin{table}[t]
\footnotesize
\centering
\caption{\textbf{Comparison of single identity unlearning methods}. Results are reported for iResNet100 FR models in terms of open-set and closed-set unlearning and face verification performance. The proposed methods show a strong reduction in unlearning time.}
\label{tab:single_identity_comparison}
\setlength{\tabcolsep}{2.5pt}
\renewcommand{\arraystretch}{0.45}
\begin{tabular}{@{}llccccccccccccc@{}}
\toprule
 &  & \multicolumn{2}{c}{Closed-set} & \multicolumn{2}{c}{Open-set} &  \\ 
 \cmidrule(lr){3-4} \cmidrule(lr){5-6} 
 {Model} & Method & Fgt. \% & Ret. \% & FS & $\Delta \text{Avg}_5$ & Time (s) \\ 
\midrule

\multirow{4}{*}{ArcFace}
& LFSSD-E   & 100.00 & 99.42 & 0.998 & 96.78 & 23009.37  \\
& RAD       & 100.00 & 99.40 & 0.968 & 96.83  & 12123.29 \\
& IS$^3$ (ours)   & 100.00 & 98.87 & 0.919 & 96.80 & 0.17 \\
& RIS (ours)       & 0.00 & 98.86 & 1.000 & 96.74 & 0.23 \\
\midrule

\multirow{4}{*}{AdaFace}
& LFSSD-E   & 100.00 & 99.43 & 0.956 & 96.73 & 23988.99 \\
& RAD       & 100.00 & 99.44 & 0.915 & 96.73 & 13874.75 \\
& IS$^3$ (ours)    & 100.00 & 98.96 & 0.735 & 96.98 & 0.23 \\
& RIS (ours)       & 0.00 & 98.97 & 1.000 & 97.01 & 0.27  \\
\midrule

\multirow{4}{*}{MagFace}
& LFSSD-E   & 100.00 & 99.47 & 1.000 & 96.85 & 22138.31 \\
& RAD       & 100.00 & 99.46 & 0.875 & 96.83 & 9074.24  \\
& IS$^3$ (ours)    & 100.00 & 98.99 & 0.990 & 97.17 & 0.15 \\
& RIS (ours)       & 0.00 & 98.99 & 1.000 & 97.15 & 0.29 \\
\midrule
\multirow{4}{*}{DQMFace}
& LFSSD-E   & 100.00 & 99.50 & 0.995 & 97.13 & 22219.67  \\
& RAD       & 100.00 & 99.48 & 0.923 & 97.19 & 9101.15 \\
& IS$^3$ (ours)    & 100.00 & 99.06 & 0.801 & 97.25 & 0.20 \\
& RIS (ours)       & 0.00 & 99.06 & 1.000 & 97.19 & 0.23 \\
\bottomrule
\end{tabular}
\end{table}

\begin{table*}[t]
\caption{\textbf{Unlearning performance evaluation}.
Results are reported for iResNet100 FR models across varying numbers of forgotten identities, considering both closed-set and open-set unlearning efficiency, and change in recognition performance. ($\Delta\mathrm{Avg}_5$) reports the change in mean verification accuracy across five benchmarks as percentage points relative to the corresponding Lin and Base performance.}
\label{tab:r100_unlearning}
\footnotesize
\centering
\setlength{\tabcolsep}{4pt}
\renewcommand{\arraystretch}{0.4}
\begin{tabular}{@{}clccccccccccccc@{}}
\toprule
 &  & \multicolumn{5}{c}{IS$^3$} & \multicolumn{5}{c}{RIS} \\ \cmidrule(lr){3-7} \cmidrule(l){8-12} 
 &  & \multicolumn{2}{c}{Closed-set} & \multicolumn{2}{c}{Open-set} &  & \multicolumn{2}{c}{Closed-set} & \multicolumn{2}{c}{Open-set} &  \\ 
 \cmidrule(lr){3-4} \cmidrule(lr){5-6} \cmidrule(lr){8-9} \cmidrule(lr){10-11} 
 {Model} & $|\mathcal{C}_{f}|$ & Fgt. \% & Ret. \% & FS & $\Delta \text{Avg}_5$ & Time (s) & Fgt. \% & Ret. \% & FS & $\Delta \text{Avg}_5$ & Time (s) \\ 
\midrule
\multirow{13}{*}{\rotatebox[origin=c]{90}{\strut ArcFace}}
& \textit{Base} & 100.00 & 99.42 & 0.000 & - & - & 100.00 & 99.42 & 0.000 & - & - \\
& \textit{Lin} & 100.00 & 98.88 & 0.000 & - & - & 100.00 & 98.88 & 0.000 & - & - \\
& 1 & 100.00 & 98.87 & 0.919 & +0.00 & 0.17 & 0.00 & 98.86 & 1.000 & -0.06 & 0.23 \\
& 5 & 16.85 & 98.88 & 1.000 & +0.00 & 0.18 & 0.00 & 98.87 & 1.000 & -0.07 & 0.25 \\
& 10 & 0.53 & 98.86 & 1.000 & -0.08 & 0.19 & 0.00 & 98.88 & 1.000 & -0.18 & 0.26 \\
& 50 & 0.64 & 98.86 & 0.991 & -0.16 & 0.20 & 0.00 & 98.72 & 1.000 & -1.36 & 0.26 \\
& 100 & 0.58 & 98.67 & 0.995 & -0.62 & 0.21 & 0.16 & 98.40 & 0.998 & -5.52 & 0.27 \\
& 500 & 54.20 & 98.29 & 0.879 & -1.75 & 0.22 & 16.66 & 94.33 & 0.824 & -9.33 & 0.29 \\
& 1K & 93.64 & 98.30 & 0.665 & -1.91 & 0.23 & 30.02 & 87.99 & 0.266 & -9.02 & 0.25 \\ 
\midrule
\multirow{13}{*}{\rotatebox[origin=c]{90}{\strut AdaFace}}
& \textit{Base} & 100.00 & 99.43 & 0.000 & - & - & 100.00 & 99.43 & 0.000 & - & - \\
& \textit{Lin} & 100.00 & 98.97 & 0.000 & - & - & 100.00 & 98.97 & 0.000 & - & - \\
& 1 & 100.00 & 98.96 & 0.735 & -0.00 & 0.23 & 0.00 & 98.93 & 1.000 & +0.03 & 0.28 \\
& 5 & 20.22 & 98.93 & 1.000 & -0.02 & 0.19 & 0.00 & 98.92 & 1.000 & -0.07 & 0.28 \\
& 10 & 1.59 & 98.94 & 1.000 & -0.12 & 0.21 & 0.00 & 98.90 & 1.000 & -0.19 & 0.27 \\
& 50 & 3.22 & 98.89 & 0.977 & -0.14 & 0.26 & 0.64 & 98.77 & 0.997 & -0.80 & 0.28 \\
& 100 & 1.62 & 98.57 & 0.978 & -0.74 & 0.23 & 3.50 & 98.71 & 0.997 & -1.03 & 0.30 \\
& 500 & 25.26 & 90.03 & 0.546 & -2.30 & 0.26 & 96.86 & 98.68 & 0.946 & -0.93 & 0.32 \\
& 1K & 62.43 & 90.44 & 0.325 & -2.56 & 0.29 & 98.20 & 98.77 & 0.842 & -0.90 & 0.33 \\ 
\midrule
\multirow{13}{*}{\rotatebox[origin=c]{90}{\strut MagFace}}
& \textit{Base} & 100.00 & 99.47 & 0.000 & - & - & 100.00 & 99.47 & 0.000 & - & - \\
& \textit{Lin} & 100.00 & 99.00 & 0.000 & - & - & 100.00 & 99.00 & 0.000 & - & - \\
& 1 & 100.00 & 98.99 & 0.990 & -0.02 & 0.15 & 0.00 & 98.99 & 1.000 & -0.04 & 0.29 \\
& 5 & 5.62 & 98.99 & 1.000 & -0.07 & 0.17 & 0.00 & 98.98 & 1.000 & -0.16 & 0.29 \\
& 10 & 0.00 & 98.99 & 1.000 & -0.16 & 0.17 & 0.00 & 98.97 & 1.000 & -0.26 & 0.29 \\
& 50 & 0.11 & 98.95 & 0.992 & -0.29 & 0.18 & 0.11 & 98.80 & 1.000 & -0.91 & 0.28 \\
& 100 & 0.00 & 97.89 & 0.991 & -1.92 & 0.19 & 1.88 & 98.45 & 0.994 & -1.25 & 0.29 \\
& 500 & 0.00 & 0.00 & 0.004 & -9.98 & 0.20 & 38.53 & 92.94 & 0.658 & -1.03 & 0.32 \\
& 1K & 0.00 & 0.00 & 0.000 & -14.95 & 0.22 & 51.64 & 86.79 & 0.438 & -0.86 & 0.33 \\ 
\midrule
\multirow{13}{*}{\rotatebox[origin=c]{90}{\strut QMagFace}}
& \textit{Base} & 100.00 & 99.47 & 0.000 & - & - & 100.00 & 99.47 & 0.000 & - & - \\
& \textit{Lin} & 100.00 & 99.00 & 0.000 & - & - & 100.00 & 99.00 & 0.000 & - & - \\
& 1 & 100.00 & 98.99 & 0.990 & -0.04 & 0.15 & 0.00 & 98.99 & 1.000 & -0.02 & 0.29 \\
& 5 & 5.62 & 98.99 & 1.000 & -0.02 & 0.17 & 0.00 & 98.98 & 1.000 & -0.17 & 0.29 \\
& 10 & 0.00 & 98.99 & 1.000 & -0.15 & 0.17 & 0.00 & 98.97 & 1.000 & -0.21 & 0.29 \\
& 50 & 0.11 & 98.95 & 0.992 & -0.22 & 0.18 & 0.11 & 98.80 & 1.000 & -0.81 & 0.28 \\
& 100 & 0.00 & 97.89 & 0.991 & -1.75 & 0.19 & 1.88 & 98.45 & 0.994 & -1.48 & 0.29 \\
& 500 & 0.00 & 0.00 & 0.004 & -2.09 & 0.20 & 38.53 & 92.94 & 0.658 & -1.24 & 0.32 \\
& 1K & 0.00 & 0.00 & 0.000 & -2.15 & 0.22 & 51.64 & 86.79 & 0.438 & -1.04 & 0.33 \\ 
\midrule
\multirow{13}{*}{\rotatebox[origin=c]{90}{\strut DQMFace}}
& \textit{Base} & 100.00 & 99.49  & 0.000 & - & - & 100.00 & 99.49 & 0.000 & - & - \\
& \textit{Lin} & 100.00 & 99.07 & 0.000 & - & - & 100.00 & 99.07 & 0.000 & - & - \\
& 1 & 100.00 & 99.06 & 0.801 & +0.03 & 0.20 & 0.00 & 99.06 & 1.000 & -0.03 & 0.23 \\
& 5 & 8.99 & 99.07 & 1.000 & -0.02 & 0.17 & 0.00 & 99.06 & 1.000 & -0.02 & 0.30 \\
& 10 & 0.00 & 99.04 & 1.000 & -0.02 & 0.17 & 0.53 & 99.03 & 1.000 & -0.10 & 0.31 \\
& 50 & 0.32 & 99.02 & 0.998 & -0.23 & 0.23 & 0.54 & 98.97 & 1.000 & -0.69 & 0.31 \\
& 100 & 0.05 & 98.70 & 0.995 & -1.04 & 0.18 & 2.04 & 98.77 & 0.997 & -0.93 & 0.26 \\
& 500 & 0.31 & 3.16 & 0.182 & -14.36 & 0.19 & 68.75 & 98.28 & 0.834 & -1.32 & 0.27 \\
& 1K & 0.72 & 3.36 & 0.129 & -15.25 & 0.21 & 83.91 & 97.98 & 0.611 & -1.03 & 0.25 \\ 
\bottomrule
\end{tabular}
\end{table*}

The proposed linearizer framework and its two closed-form identity-level unlearning methods, IS³ and RIS, are evaluated across different FR models and benchmarks. The evaluation focuses on examining the unlearning effectiveness in closed-set versus open-set settings, scalability of unlearning varying numbers of identities, preservation of face verification performance, and computational efficiency relative to existing unlearning approaches.


\subsection{Linearizer Fidelity}
Before applying identity unlearning, the impact of the linearizer framework on the original FR performance is assessed across all evaluated models. As shown in Tables \ref{tab:r100_unlearning},  \ref{tab:r100_ver_benchmarks}, and \ref{tab:r100_ijb_ver}. the iResNet100 linearizer (Lin) preserves the recognition performance of the original models, with only negligible differences in verification accuracy and TAR. These results indicate that the discriminative capabilities of the pretrained FR models are maintained after linearization, providing a reliable foundation for subsequent identity unlearning through the linear operations.

\subsection{Identity-Level Unlearning Effectiveness}
\begin{table*}[!h]
\footnotesize
\caption{\textbf{Impact of unlearning on small-scale face verification performance.} 
Results are reported for iResNet100 FR models across five verification benchmarks, showing changes in mean accuracy after unlearning relative to the corresponding linearized model (Lin) performance. Changes are expressed in percentage points, with negative values indicating performance degradation.}
\label{tab:r100_ver_benchmarks}
\centering
\setlength{\tabcolsep}{4pt}
\renewcommand{\arraystretch}{0.3}
\begin{tabular}{@{}clcccccccccc@{}}
\toprule
& & \multicolumn{5}{c}{IS$^3$}
    & \multicolumn{5}{c}{RIS} \\
\cmidrule(lr){3-7}
\cmidrule(l){8-12}
Model & $|\mathcal{C}_{f}|$
& LFW & CFP-FP & AgeDB & CALFW & CPLFW
& LFW & CFP-FP & AgeDB & CALFW & CPLFW \\
\midrule
\multirow{15}{*}{\rotatebox[origin=c]{90}{\strut ArcFace}}
& \textit{Base} & 99.78 & 97.90 & 97.89 & 95.97 & 92.43 & 99.78 & 97.90 & 97.89 & 95.97 & 92.43 \\
& \textit{Lin} & 99.77 & 97.87 & 97.92 & 95.97 & 92.45 & 99.77 & 97.87 & 97.92 & 95.97 & 92.45 \\
& 1 & +0.00 & +0.06 & +0.07 & +0.03 & -0.13 & +0.02 & -0.07 & -0.05 & +0.02 & -0.20 \\
& 5 & -0.05 & +0.01 & +0.03 & -0.02 & +0.00 & -0.03 & -0.26 & -0.18 & +0.08 & +0.05 \\
& 10 & -0.03 & -0.11 & +0.00 & +0.08 & -0.32 & -0.03 & -0.27 & -0.28 & +0.00 & -0.32 \\
& 50 & -0.02 & -0.06 & -0.30 & +0.07 & -0.48 & -0.05 & -2.17 & -1.42 & -0.42 & -2.75 \\
& 100 & -0.03 & -1.00 & -0.90 & -0.20 & -0.95 & -0.88 & -8.29 & -6.55 & -3.88 & -7.98 \\
& 500 & -0.30 & -2.64 & -2.15 & -1.15 & -2.48 & -2.97 & -13.43 & -12.82 & -6.62 & -10.83 \\
& 1K & -0.27 & -3.17 & -2.52 & -1.13 & -2.45 & -3.23 & -13.19 & -11.53 & -6.75 & -10.42 \\
\midrule
\multirow{15}{*}{\rotatebox[origin=c]{90}{\strut AdaFace}}
& \textit{Base} & 99.75 & 98.13 & 98.19 & 96.01 & 92.82 & 99.75 & 98.13 & 98.19 & 96.01 & 92.82 \\
& \textit{Lin} & 99.77 & 98.19 & 98.08 & 96.00 & 92.85 & 99.77 & 98.19 & 98.08 & 96.00 & 92.85 \\
& 1 & -0.02 & +0.06 & -0.08 & +0.05 & -0.02 & +0.02 & +0.06 & +0.08 & -0.03 & +0.02 \\
& 5 & -0.03 & -0.01 & -0.10 & +0.17 & -0.13 & +0.03 & -0.10 & -0.35 & -0.05 & +0.10 \\
& 10 & -0.03 & -0.13 & -0.13 & -0.02 & -0.27 & +0.00 & -0.47 & -0.28 & +0.05 & -0.25 \\
& 50 & 0.00 & -0.37 & -0.30 & +0.03 & -0.07 & -0.08 & -1.43 & -1.32 & -0.23 & -0.93 \\
& 100 & +0.02 & -1.23 & -1.17 & -0.12 & -1.18 & -0.22 & -1.50 & -1.42 & -0.53 & -1.48 \\
& 500 & -0.33 & -3.57 & -3.62 & -1.63 & -2.35 & -0.15 & -2.06 & -0.80 & -0.47 & -1.20 \\
& 1K & -0.30 & -3.50 & -4.37 & -1.55 & -3.08 & -0.12 & -1.86 & -0.93 & -0.32 & -1.30 \\
\midrule
\multirow{15}{*}{\rotatebox[origin=c]{90}{\strut MagFace}}
& \textit{Base} & 99.80 & 98.38 & 98.30 & 96.08 & 92.83 & 99.80 & 98.38 & 98.30 & 96.08 & 92.83 \\
& \textit{Lin} & 99.82 & 98.43 & 98.25 & 96.15 & 93.28 & 99.82 & 98.43 & 98.25 & 96.15 & 93.28 \\
& 1 & +0.00 & +0.04 & -0.08 & +0.02 & -0.08 & +0.00 & -0.13 & -0.03 & +0.00 & -0.03 \\
& 5 & +0.00 & -0.01 & -0.12 & -0.01 & -0.12 & -0.02 & -0.26 & -0.27 & -0.13 & -0.15 \\
& 10 & -0.03 & -0.09 & -0.17 & -0.22 & -0.32 & +0.00 & -0.27 & -0.42 & -0.18 & -0.45 \\
& 50 & -0.05 & -0.3 & -0.42 & -0.15 & -0.53 & -0.13 & -1.40 & -1.05 & -0.47 & -1.48 \\
& 100 & -0.02 & -2.57 & -2.73 & -1.10 & -3.02 & -0.08 & -1.79 & -1.47 & -0.83 & -2.08 \\
& 500 & -6.22 & -11.2 & -13.5 & -10.63 & -8.35 & -0.03 & -1.51 & -1.35 & -0.73 & -1.52 \\
& 1K & -5.07 & -17.54 & -21.38 & -18.6 & -12.17 & -0.03 & -0.33 & +0.15 & -0.20 & -0.30 \\
\midrule
\multirow{15}{*}{\rotatebox[origin=c]{90}{\strut QMagFace}}
& \textit{Base} & 99.80 & 98.23 & 98.40 & 96.10 & 93.85 &  99.80 & 98.23 & 98.40 & 96.10 & 93.85 \\
& \textit{Lin} & 99.83 & 98.74 & 98.32 & 96.03 & 94.00 & 99.83 & 98.74 & 98.32 & 96.03 & 94.00 \\
& 1 & +0.00 & -0.07 & -0.07 & -0.03 & -0.07 & -0.03 & -0.26 & -0.07 & +0.22 & +0.03 \\
& 5 & -0.03 & -0.11 & -0.02 & -0.05 & +0.13 & -0.07 & -0.37 & -0.20 & -0.03 & -0.20 \\
& 10 & -0.08 & -0.13 & -0.20 & -0.15 & -0.17 & -0.07 & -0.40 & -0.40 & -0.08 & -0.12 \\
& 50 & -0.02 & -0.36 & -0.35 & -0.12 & -0.28 & -0.02 & -1.27 & -1.03 & -0.25 & -1.48 \\
& 100 & -0.18 & -2.51 & -2.42 & -0.80 & -2.83 & -0.08 & -2.01 & -1.50 & -0.85 & -2.93 \\
& 500 & -0.08 & -2.63 & -2.70 & -0.78 & -4.23 & -0.05 & -2.00 & -1.22 & -0.70 & -2.25 \\
& 1K & -0.15 & -2.80 & -2.77 & -0.98 & -4.03 & -0.08 & -1.59 & -0.90 & -0.47 & -2.17 \\
\midrule
\multirow{15}{*}{\rotatebox[origin=c]{90}{\strut DQMFace}}
& \textit{Base} & 99.82 & 98.53 & 98.28 & 96.05 & 93.05 & 99.82 & 98.53 & 98.28 & 96.05 & 93.05 \\   
& \textit{Lin} & 99.82 & 98.79 & 98.17 & 96.02 & 93.30 & 99.82 & 98.79 & 98.17 & 96.02 & 93.30 \\
& 1 & +0.00 & +0.04 & +0.05 & +0.07 & -0.02 & -0.02 & -0.04 & -0.02 & +0.02 & -0.10 \\
& 5 & -0.05 & +0.04 & -0.02 & +0.08 & -0.05 & -0.02 & -0.10 & +0.12 & +0.08 & -0.20 \\
& 10 & -0.03 & -0.03 & +0.12 & +0.08 & -0.22 & +0.02 & -0.30 & -0.05 & +0.17 & -0.33 \\
& 50 & -0.02 & -0.63 & -0.15 & +0.03 & -0.37 & -0.13 & -1.07 & -0.75 & -0.30 & -1.18 \\
& 100 & -0.08 & -1.53 & -1.28 & -0.65 & -1.65 & -0.07 & -1.44 & -0.93 & -0.37 & -1.85 \\
& 500 & -6.47 & -16.79 & -20.88 & -13.45 & -14.22 & -0.20 & -2.41 & -1.33 & -0.53 & -2.13 \\
& 1K & -7.15 & -17.90 & -21.67 & -14.22 & -15.30 & -0.10 & -1.80 & -0.92 & -0.55 & -1.80 \\
\bottomrule
\end{tabular}
\end{table*}

The unlearning effectiveness of the closed-form strategies IS$^3$ and RIS using the open-set FS and the closed-set forget accuracy are reported in Table \ref{tab:r100_unlearning}. While forget accuracy provides a complementary evaluation aligned with existing machine unlearning research, it doesn't necessarily reflect the extent of forgetting in the embedding space. 

The two methods exhibit distinct forgetting behaviors. For single-identity unlearning, IS$^3$ maintains a forget accuracy of 100\% across all evaluated models, despite achieving FS values ranging from 0.735 to 0.990. This behavior can be attributed to its subspace-based suppression mechanism, where removing identity-related directions can significantly alter the embeddings without changing their classification outcomes. As $|\mathcal{C}_f|$ increases, the estimated subspace captures a broader range of identity-related directions, corresponding with near-zero forget accuracy for most models at 10-100 identities, while maintaining FS values close to 1.000. 
In contrast, RIS achieves zero forget accuracy and an FS of one for single-identity unlearning across all evaluated models. This can be explained by its explicit mapping of forget representations toward a target, shifting them away from their original identity-specific representations. This targeted modification provides greater control over the resulting embeddings than subspace suppression alone, potentially explaining its stronger closed-set forgetting at smaller identity scales while maintaining high FS scores. These observations highlight the discrepancy between classification-based unlearning and changes in the learned identity representations, emphasizing the importance of open-set evaluation for FR unlearning.

\subsection{Scalability Across Forget Identity Sets}
As described in Section \ref{sec:setup}, unlearning effectiveness is evaluated for different values of $\mathcal{C}_{f}$. The scalability of IS$^3$ and RIS is analyzed by increasing the size of $\mathcal{C}_f$ from a single identity to 1K as reported in Table \ref{tab:r100_unlearning}. Both methods demonstrate strong open-set forgetting effectiveness for up to 100 identities, consistently achieving FS scores close to one across all evaluated FR models. However, their effectiveness declines at larger $|\mathcal{C}_{f}|$, with distinct patterns across methods and architectures.
For IS$^3$, the decline becomes noticeable at 500 and 1K identities, especially for MagFace, QMagFace, and DQMFace. This behavior may be associated with the dimensionality of the suppressed identity subspace, controlled by the hyperparameter $k$. For these scales, the tuned value of $k=256$ corresponds to a subspace spanning $50\%$ of the latent dimensions, potentially affecting directions shared with $\mathcal{D}_{r}$. Hence, modifications to $A$ become less selective, reducing the separation between forget and retain representation changes measured by FS. Meanwhile, RIS maintains higher FS values at these scales, suggesting greater robustness to increasing unlearning requests. Its regularization of the retain representations may help limit unintended modifications, although its unlearning effectiveness also deteriorates as $|\mathcal{C}_{f}|$ increases.
\subsection{Face Verification Performance}
\begin{table*}[t]
\footnotesize
\caption{\textbf{Impact of unlearning on large-scale face verification performance.}
Results are reported for iResNet100 FR models on IJB-B and IJB-C using TAR at FAR thresholds of $10^{-5}$, $10^{-4}$, and $10^{-3}$. The Base and Lin performance is reported in absolute values, while changes after unlearning are expressed in percentage points relative to the corresponding Lin, with negative values indicating degradation.}
\label{tab:r100_ijb_ver}
\centering
\setlength{\tabcolsep}{4pt}
\renewcommand{\arraystretch}{0.3}
\begin{tabular}{@{}clcccccccccccc@{}}
\toprule
& & \multicolumn{6}{c}{IS$^3$} & \multicolumn{6}{c}{RIS} \\
\cmidrule(lr){3-8} \cmidrule(l){9-14}
& & \multicolumn{3}{c}{IJB-B} & \multicolumn{3}{c}{IJB-C}
& \multicolumn{3}{c}{IJB-B} & \multicolumn{3}{c}{IJB-C} \\
\cmidrule(lr){3-5} \cmidrule(lr){6-8} \cmidrule(lr){9-11} \cmidrule(l){12-14}
Model & $|\mathcal{C}_f|$ & $10^{-5}$ & $10^{-4}$ & $10^{-3}$ & $10^{-5}$ & $10^{-4}$ & $10^{-3}$
& $10^{-5}$ & $10^{-4}$ & $10^{-3}$ & $10^{-5}$ & $10^{-4}$ & $10^{-3}$ \\
\midrule
\multirow{15}{*}{\rotatebox[origin=c]{90}{\strut ArcFace}}
& \textit{Base} & 86.35 & 94.63 & 96.52 & 93.14 & 96.12 & 97.46 & 86.35 & 94.63 & 96.52 & 93.14 & 96.12 & 97.46 \\
& \textit{Lin} & 85.98 & 94.35 & 96.48 & 92.43 & 95.96 & 97.40 & 85.98 & 94.35 & 96.48 & 92.43 & 95.96 & 97.40 \\
& 1 & +0.11 & -0.05 & +0.04 & -0.04 & +0.01 & +0.01 & +0.35 & -0.09 & -0.05 & +0.09 & -0.07 & +0.00 \\
& 5 & -0.16 & -0.07 & +0.00 & -0.01 & -0.00 & +0.02 & -0.06 & -0.40 & -0.17 & -0.31 & -0.24 & -0.14 \\
& 10 & -0.27 & -0.19 & +0.03 & -0.19 & -0.11 & -0.04 & +0.41 & -0.35 & -0.30 & -0.27 & -0.36 & -0.26 \\
& 50 & -0.33 & -0.33 & -0.12 & -0.72 & -0.26 & -0.10 & -1.62 & -2.77 & -2.01 & -2.08 & -2.15 & -1.67 \\
& 100 & -0.52 & -1.33 & -0.75 & -1.39 & -1.04 & -0.64 & -18.01 & -17.54 & -12.85 & -21.87 & -16.24 & -10.58 \\
& 500 & -2.94 & -3.04 & -2.04 & -3.73 & -2.51 & -1.61 & -34.79 & -32.07 & -23.68 & -30.85 & -25.77 & -18.60 \\
& 1K & -3.46 & -2.94 & -2.23 & -3.58 & -2.47 & -1.75 & -32.65 & -30.14 & -22.83 & -28.06 & -23.89 & -18.20 \\
\midrule
\multirow{15}{*}{\rotatebox[origin=c]{90}{\strut AdaFace}}
& \textit{Base} & 89.07 & 95.21 & 96.82 & 94.34 & 96.50 & 97.70 & 89.07 & 95.21 & 96.82 & 94.34 & 96.50 & 97.70 \\
& \textit{Lin} & 89.08 & 95.20 & 96.81 & 94.32 & 96.50 & 97.69 & 89.08 & 95.20 & 96.81 & 94.32 & 96.50 & 97.69 \\
& 1 & -0.34 & -0.02 & +0.00 & +0.05 & -0.02 & +0.02 & -0.41 & -0.07 & -0.04 & -0.02 & -0.02 & -0.02 \\
& 5 & -0.49 & -0.02 & -0.02 & +0.02 & -0.01 & +0.00 & -0.61 & -0.20 & -0.12 & -0.36 & -0.15 & -0.10 \\
& 10 & -0.97 & -0.14 & -0.14 & -0.18 & -0.16 & -0.05 & -1.25 & -0.47 & -0.37 & -0.65 & -0.33 & -0.32 \\
& 50 & -1.38 & -0.30 & -0.29 & -0.46 & -0.26 & -0.22 & -1.07 & -1.95 & -1.20 & -2.09 & -1.59 & -1.07 \\
& 100 & -0.59 & -1.42 & -0.90 & -1.58 & -1.11 & -0.73 & -1.85 & -2.60 & -1.85 & -2.46 & -1.96 & -1.46 \\
& 500 & -8.29 & -6.78 & -4.24 & -7.53 & -4.90 & -3.24 & -8.31 & -7.57 & -4.87 & -6.83 & -4.92 & -3.46 \\
& 1K & -10.18 & -7.24 & -4.51 & -7.95 & -5.37 & -3.54 & -10.95 & -9.74 & -6.09 & -8.59 & -6.39 & -4.16 \\
\midrule
\multirow{15}{*}{\rotatebox[origin=c]{90}{\strut MagFace}}
& \textit{Base} & 88.09 & 94.25 & 96.50 & 92.40 & 95.68 & 97.35 & 88.09 & 94.25 & 96.50 & 92.40 & 95.68 & 97.35 \\
& \textit{Lin} & 90.55 & 94.89 & 96.55 & 94.05 & 96.14 & 97.53 & 90.55 & 94.89 & 96.55 & 94.05 & 96.14 & 97.53 \\
& 1 & +0.00 & +0.01 & +0.03 & +0.01 & +0.04 & -0.01 & +0.50 & -0.06 & +0.02 & -0.04 & +0.00 & -0.02 \\
& 5 & +0.24 & -0.07 & +0.02 & +0.16 & +0.03 & -0.02 & +0.07 & -0.23 & -0.08 & -0.11 & -0.11 & -0.12 \\
& 10 & +0.23 & -0.12 & -0.09 & +0.01 & +0.02 & -0.09 & -0.42 & -0.45 & -0.44 & -0.41 & -0.33 & -0.42 \\
& 50 & -1.61 & -0.61 & -0.33 & -0.61 & -0.40 & -0.30 & -3.53 & -2.40 & -1.67 & -2.40 & -1.67 & -1.32 \\
& 100 & -5.28 & -4.17 & -2.88 & -4.92 & -3.26 & -2.35 & -12.59 & -6.91 & -4.43 & -8.33 & -5.31 & -3.38 \\
& 500 & -25.98 & -22.1 & -15.63 & -20.59 & -16.13 & -11.61 & -22.47 & -13.64 & -8.14 & -12.90 & -8.85 & -5.56 \\ 
& 1K & -27.08 & -22.37 & -16.17 & -20.90 & -16.53 & -11.95 & -21.25 & -13.9 & -8.49 & -12.75 & -9.01 & -5.79 \\
\midrule
\multirow{15}{*}{\rotatebox[origin=c]{90}{\strut QMagFace}}
& \textit{Base} & 90.19 & 94.71 & 96.32 & 93.75 & 96.17 & 97.56 & 90.19 & 94.71 & 96.32 & 93.75 & 96.17 & 97.56 \\
& \textit{Lin} & 90.61 & 94.94 & 96.41 & 94.01 & 96.40 & 97.67 & 90.61 & 94.94 & 96.41 & 94.01 & 96.40 & 97.67 \\
& 1 & -0.05 & +0.01 & +0.06 & +0.01 & +0.00 & +0.01 & +0.04 & +0.01 & +0.06 & -0.03 & -0.01 & -0.02 \\
& 5 & +0.02 & +0.02 & +0.07 & +0.05 & +0.05 & -0.04 & -0.22 & -0.15 & +0.02 & -0.19 & -0.15 & -0.10 \\
& 10 & +0.12 & -0.03 & +0.03 & +0.00 & -0.03 & -0.06 & -0.41 & -0.42 & -0.12 & -0.35 & -0.35 & -0.27 \\
& 50 & -1.05 & -0.65 & -0.15 & -0.49 & -0.39 & -0.22 & -2.84 & -1.99 & -1.26 & -1.87 & -1.60 & -1.14 \\
& 100 & -4.71 & -3.64 & -2.05 & -4.14 & -2.98 & -1.93 & -9.38 & -5.80 & -3.63 & -6.79 & -4.63 & -3.05 \\
& 500 & -24.11 & -19.56 & -13.21 & -18.07 & -14.30 & -9.85 & -18.86 & -11.39 & -6.68 & -10.35 & -7.43 & -4.76 \\
& 1K & -24.89 & -19.80 & -13.48 & -18.89 & -14.65 & -10.09 & -18.08 & -11.36 & -7.07 & -10.48 & -7.35 & -4.84 \\
\midrule
\multirow{15}{*}{\rotatebox[origin=c]{90}{\strut DQMFace}}
& \textit{Base} & 91.35 & 95.38 & 96.71 & 94.84 & 96.58 & 97.65 & 91.35 & 95.38 & 96.71 & 94.84 & 96.58 & 97.65 \\
& \textit{Lin} & 90.61 & 95.10 & 96.64 & 94.57 & 96.45 & 97.62 & 90.61 & 95.10 & 96.64 & 94.57 & 96.45 & 97.62 \\
& 1 & +0.04 & -0.01 & +0.02 & +0.03 & -0.01 & -0.01 & +0.27 & -0.02 & -0.04 & -0.08 & -0.01 & -0.07 \\
& 5 & +0.15 & -0.05 & -0.02 & +0.02 & -0.02 & -0.01 & -0.04 & -0.28 & -0.19 & -0.30 & -0.19 & -0.17 \\
& 10 & -0.06 & -0.18 & -0.14 & +0.03 & -0.10 & -0.13 & -0.75 & -0.48 & -0.42 & -0.51 & -0.38 & -0.32 \\
& 50 & -0.35 & -0.34 & -1.88 & -0.33 & -0.24 & -1.25 & -1.90 & -1.58 & -1.01 & -1.96 & -1.22 & -0.85 \\
& 100 & -2.67 & -2.30 & -1.53 & -2.75 & -1.88 & -1.25 & -5.45 & -4.55 & -3.02 & -4.99 & -3.46 & -2.24 \\
& 500 & -56.78 & -45.10 & -35.07 & -55.97 & -45.10 & -31.81 & -18.41 & -13.17 & -7.9 & -11.7 & -8.34 & -5.31 \\
& 1K & -57.91 & -49.61 & -36.51 & -57.54 & -47.14 & -33.71 & -18.54 & -13.44 & -8.14 & -12.15 & -8.51 & -5.55 \\
\bottomrule
\end{tabular}
\end{table*}

Tables \ref{tab:r100_ver_benchmarks} and \ref{tab:r100_ijb_ver} show that both IS$^3$ and RIS largely preserve baseline verification up to 50 unlearned identities, while recognition declines at larger $|\mathcal{C}_{f}|$, reflecting how differently each method suppresses identity information.
As shown in Table \ref{tab:r100_ver_benchmarks}, at $\{100, 500, 1K\}$ identities, IS$^3$ degrades verification notably for MagFace and DQMFace, dropping 10 to 14 percentage points on average at 500 identities and further at 1K, coinciding with the drop in unlearning performance. The smaller degradation observed for QMagFace can be linked to its quality-aware similarity scoring during verification. RIS yields smaller reductions, except for ArcFace ($5.52$, $9.33$ and $9.02$ points at 100, 500 and 1K). This preservation is attributable to the retain regularization controlled by $\lambda$, which constrains modifications to retained representations.
On IJB-B/C, the trends are similar but degradation is more distinct, especially at stricter FAR thresholds. Up to 50 unlearned identities, both methods show limited TAR degradation, with maximum reductions of $2.75$ and $3.46$ percentage points for IS$^3$ and RIS across all models and FAR thresholds. At 500 and 1K identities, IS$^3$ degrades significantly, consistent with the reduced retain accuracy and FS in Table~\ref{tab:r100_unlearning}. At these scales IS$^3$ loses both recognition utility and selective unlearning effectiveness. RIS degrades significantly at 100, 500 and 1K identities, particularly for ArcFace, while the remaining models decline less severely, possibly because their norm-aware objectives incorporate embedding magnitude information. 

\subsection{Comparison with Existing Unlearning Methods}
Table \ref{tab:single_identity_comparison} compares IS$^3$ and RIS against some existing single-identity unlearning methods RAD and LFSSD-E. While all methods largely preserve face verification performance, their unlearning effectiveness differs. RIS achieves an FS of 1.000 and a forget accuracy of zero across all evaluated methods in less than 0.3 seconds, outperforming the compared methods. As previously observed with IS$^3$, RAD and LFSSD-E also achieve 100\% forget accuracy despite their high FS values, highlighting how closed-set metrics can lead to misleading interpretations of unlearning effectiveness in open-set settings such as FR. Beyond their unlearning effectiveness and preservation of recognition performance, both proposed methods offer a substantial computational advantage by reducing the unlearning time from several hours to fractions of a second. 

\section{Conclusion}
Prior machine unlearning methods largely assume closed-set classification, where unlearning amounts to removing a discrete output class from a small-scale model. This formulation does not transfer to FR, because identity information is spread nonlinearly across the embedding and must be removed under open-set evaluation. We introduced a linearizer framework for identity unlearning that maps face embeddings into a space where identity information can be modified by learnable linear operations, and proposed two closed-form strategies on top of it, IS$^3$ and RIS.
Both strategies avoid retraining entirely. A removal request is served by a closed-form computation over the linearized space, so single and simultaneous multi-identity deletions complete in milliseconds. Experiments across architectures, objectives, and datasets show selective, effective unlearning with largely preserved recognition performance, though preservation degrades as the number of forgotten identities grows.

\clearpage
\newpage

\section*{LLM Usage Statement}
The authors used locally hosted DeepSeek V4.1 to correct the grammar, proofread the manuscript, and improve its clarity and readability. The authors reviewed and verified all AI-assisted revisions and take full responsibility for the content of this work.

{
    \small
    \bibliographystyle{ieeenat_fullname}
    \bibliography{main}

@String(CVPR= {IEEE Conf. Comput. Vis. Pattern Recog.})

@String(ECCV= {Eur. Conf. Comput. Vis.})

@String(ICLR = {Int. Conf. Learn. Represent.})

@String(AAAI = {AAAI})

@String(CVPR  = {CVPR})

@String(ECCV  = {ECCV})

@String(ICLR  = {ICLR})

@misc{alrefai2026unlearningspeedrapidactivation,
      title={Unlearning at Speed: Rapid Activation Dampening for Efficient Machine Unlearning in Real-World Face Recognition}, 
      author={Rouqaiah Al-Refai and Leo Scheubel and Philipp Terhörst},
      year={2026},
      eprint={2611.11433},
      archivePrefix={arXiv},
      primaryClass={cs.CV},
      url={https://arxiv.org/abs/2611.11433}, 
}

@misc{berman2026linearizer,
  title        = {Who Said Neural Networks Aren't Linear?},
  author       = {Nimrod Berman and Assaf Hallak and Assaf Shocher},
  journal      = {International Conference on Machine Learning (ICML)},
  year         = 2026
}

@INPROCEEDINGS{gandikota2024unified,
  author={Gandikota, Rohit and Orgad, Hadas and Belinkov, Yonatan and Materzyńska, Joanna and Bau, David},
  booktitle={2024 IEEE/CVF Winter Conference on Applications of Computer Vision (WACV)}, 
  title={Unified Concept Editing in Diffusion Models}, 
  year={2024},
  volume={},
  number={},
  pages={5099-5108},
  doi={10.1109/WACV57701.2024.00503}}

@misc{eu_gdpr_2016,
  title        = {{Regulation (EU) 2016/679 of the European Parliament and of the Council of 27 April 2016 on the protection of natural persons with regard to the processing of personal data and on the free movement of such data (General Data Protection Regulation)}},
  author       = {{European Union}},
  year         = {2016},
  howpublished = {\url{https://eur-lex.europa.eu/eli/reg/2016/679/oj}},
  note         = {Official Journal of the European Union, L 119, 1--88},
}

@ARTICLE{9481149privacy,
  author={Meden, Blaž and Rot, Peter and Terhörst, Philipp and Damer, Naser and Kuijper, Arjan and Scheirer, Walter J. and Ross, Arun and Peer, Peter and Štruc, Vitomir},
  journal={IEEE Transactions on Information Forensics and Security}, 
  title={Privacy–Enhancing Face Biometrics: A Comprehensive Survey}, 
  year={2021},
  volume={16},
  number={},
  pages={4147-4183},
  doi={10.1109/TIFS.2021.3096024}
}

@INPROCEEDINGS{7163042towards,
  author={Cao, Yinzhi and Yang, Junfeng},
  booktitle={2015 IEEE Symposium on Security and Privacy}, 
  title={Towards Making Systems Forget with Machine Unlearning}, 
  year={2015},
  volume={},
  number={},
  pages={463-480},
  doi={10.1109/SP.2015.35}
}

@inproceedings{DBLP:conf/alt/Neel0S21,
  author       = {Seth Neel and
                  Aaron Roth and
                  Saeed Sharifi{-}Malvajerdi},
  editor       = {Vitaly Feldman and
                  Katrina Ligett and
                  Sivan Sabato},
  title        = {Descent-to-Delete: Gradient-Based Methods for Machine Unlearning},
  booktitle    = {Algorithmic Learning Theory, 16-19 March 2021, Virtual Conference,
                  Worldwide},
  series       = {Proceedings of Machine Learning Research},
  volume       = {132},
  pages        = {931--962},
  publisher    = {{PMLR}},
  year         = {2021},
  url          = {http://proceedings.mlr.press/v132/neel21a.html},
  bibsource    = {dblp computer science bibliography, https://dblp.org}
}

@inproceedings{DBLP:conf/nips/GuptaJNRSW21,
  author       = {Varun Gupta and
                  Christopher Jung and
                  Seth Neel and
                  Aaron Roth and
                  Saeed Sharifi{-}Malvajerdi and
                  Chris Waites},
  editor       = {Marc'Aurelio Ranzato and
                  Alina Beygelzimer and
                  Yann N. Dauphin and
                  Percy Liang and
                  Jennifer Wortman Vaughan},
  title        = {Adaptive Machine Unlearning},
  booktitle    = {Advances in Neural Information Processing Systems 34: Annual Conference
                  on Neural Information Processing Systems 2021, NeurIPS 2021, December
                  6-14, 2021, virtual},
  pages        = {16319--16330},
  year         = {2021},
  url          = {https://proceedings.neurips.cc/paper/2021/hash/87f7ee4fdb57bdfd52179947211b7ebb-Abstract.html},
  bibsource    = {dblp computer science bibliography, https://dblp.org}
}

@inproceedings{DBLP:conf/nips/SekhariAKS21,
  author       = {Ayush Sekhari and
                  Jayadev Acharya and
                  Gautam Kamath and
                  Ananda Theertha Suresh},
  editor       = {Marc'Aurelio Ranzato and
                  Alina Beygelzimer and
                  Yann N. Dauphin and
                  Percy Liang and
                  Jennifer Wortman Vaughan},
  title        = {Remember What You Want to Forget: Algorithms for Machine Unlearning},
  booktitle    = {Advances in Neural Information Processing Systems 34: Annual Conference
                  on Neural Information Processing Systems 2021, NeurIPS 2021, December
                  6-14, 2021, virtual},
  pages        = {18075--18086},
  year         = {2021},
  url          = {https://proceedings.neurips.cc/paper/2021/hash/9627c45df543c816a3ddf2d8ea686a99-Abstract.html},
  bibsource    = {dblp computer science bibliography, https://dblp.org}
}

@inproceedings{DBLP:conf/cvpr/GolatkarAS20,
  author       = {Aditya Golatkar and
                  Alessandro Achille and
                  Stefano Soatto},
  title        = {Eternal Sunshine of the Spotless Net: Selective Forgetting in Deep
                  Networks},
  booktitle    = {2020 {IEEE/CVF} Conference on Computer Vision and Pattern Recognition,
                  {CVPR} 2020, Seattle, WA, USA, June 13-19, 2020},
  pages        = {9301--9309},
  publisher    = {Computer Vision Foundation / {IEEE}},
  year         = {2020},
  url          = {https://openaccess.thecvf.com/content\_CVPR\_2020/html/Golatkar\_Eternal\_Sunshine\_of\_the\_Spotless\_Net\_Selective\_Forgetting\_in\_Deep\_CVPR\_2020\_paper.html},
  doi          = {10.1109/CVPR42600.2020.00932},
  bibsource    = {dblp computer science bibliography, https://dblp.org}
}

@inproceedings{DBLP:conf/aaai/GravesNG21,
  author       = {Laura Graves and
                  Vineel Nagisetty and
                  Vijay Ganesh},
  title        = {Amnesiac Machine Learning},
  booktitle    = {Thirty-Fifth {AAAI} Conference on Artificial Intelligence, {AAAI}
                  2021, Thirty-Third Conference on Innovative Applications of Artificial
                  Intelligence, {IAAI} 2021, The Eleventh Symposium on Educational Advances
                  in Artificial Intelligence, {EAAI} 2021, Virtual Event, February 2-9,
                  2021},
  pages        = {11516--11524},
  publisher    = {{AAAI} Press},
  year         = {2021},
  url          = {https://doi.org/10.1609/aaai.v35i13.17371},
  doi          = {10.1609/AAAI.V35I13.17371},
  bibsource    = {dblp computer science bibliography, https://dblp.org}
}

@inproceedings{DBLP:conf/nips/KurmanjiTHT23,
  author       = {Meghdad Kurmanji and
                  Peter Triantafillou and
                  Jamie Hayes and
                  Eleni Triantafillou},
  editor       = {Alice Oh and
                  Tristan Naumann and
                  Amir Globerson and
                  Kate Saenko and
                  Moritz Hardt and
                  Sergey Levine},
  title        = {Towards Unbounded Machine Unlearning},
  booktitle    = {Advances in Neural Information Processing Systems 36: Annual Conference
                  on Neural Information Processing Systems 2023, NeurIPS 2023, New Orleans,
                  LA, USA, December 10 - 16, 2023},
  year         = {2023},
  url          = {http://papers.nips.cc/paper\_files/paper/2023/hash/062d711fb777322e2152435459e6e9d9-Abstract-Conference.html},
  bibsource    = {dblp computer science bibliography, https://dblp.org}
}

@misc{hayase2020selectiveforgettingdeepnetworks,
      title={Selective Forgetting of Deep Networks at a Finer Level than Samples}, 
      author={Tomohiro Hayase and Suguru Yasutomi and Takashi Katoh},
      year={2020},
      eprint={2012.11849},
      archivePrefix={arXiv},
      primaryClass={stat.ML},
      url={https://arxiv.org/abs/2012.11849}, 
}

@Article{jcp6040121,
AUTHOR = {Nagasubramaniam, Piyush and Karamchandani, Neeraj and Wu, Chen and Zhu, Sencun},
TITLE = {Text-to-Unlearn: Robust Concept Removal in GANs via Text Prompts},
JOURNAL = {Journal of Cybersecurity and Privacy},
VOLUME = {6},
YEAR = {2026},
NUMBER = {4},
ARTICLE-NUMBER = {121},
URL = {https://www.mdpi.com/2624-800X/6/4/121},
ISSN = {2624-800X},
DOI = {10.3390/jcp6040121}
}

@inproceedings{DBLP:conf/fgr/GundogduUU24,
  author       = {Emircan G{\"{u}}ndogdu and
                  Altay Unal and
                  Gozde Unal},
  title        = {A Study Regarding Machine Unlearning on Facial Attribute Data},
  booktitle    = {18th {IEEE} International Conference on Automatic Face and Gesture
                  Recognition, {FG} 2024, Istanbul, Turkey, May 27-31, 2024},
  pages        = {1--5},
  publisher    = {{IEEE}},
  year         = {2024},
  url          = {https://doi.org/10.1109/FG59268.2024.10581972},
  doi          = {10.1109/FG59268.2024.10581972},
  bibsource    = {dblp computer science bibliography, https://dblp.org}
}

@article{Choi2023TowardsMU,
  title={Towards Machine Unlearning Benchmarks: Forgetting the Personal Identities in Facial Recognition Systems},
  author={Dasol Choi and Dongbin Na},
  journal={ArXiv},
  year={2023},
  volume={abs/2311.02240},
  url={https://api.semanticscholar.org/CorpusID:265033403}
}

@misc{tok2026nearesttargetwrongone,
      title={The Nearest Target Is the Wrong One: Target Separation in Arc2Face Identity Unlearning}, 
      author={Zeynel Tok},
      year={2026},
      eprint={2608.30087},
      archivePrefix={arXiv},
      primaryClass={cs.CV},
      url={https://arxiv.org/abs/2608.30087}, 
}

@inproceedings{DBLP:conf/pkdd/HossainMLK25,
  author       = {Prommy Sultana Hossain and
                  Emanuela Marasco and
                  Jessica Lin and
                  Michael King},
  editor       = {Rita P. Ribeiro and
                  Bernhard Pfahringer and
                  Nathalie Japkowicz and
                  Pedro Larra{\~{n}}aga and
                  Al{\'{\i}}pio M. Jorge and
                  Carlos Soares and
                  Pedro H. Abreu and
                  Jo{\~{a}}o Gama},
  title        = {"I Forgot About You": Exploring Multi-Label Unlearning {(MLU)}
                  for Responsible Facial Recognition Systems},
  booktitle    = {Machine Learning and Knowledge Discovery in Databases. Research Track
                  - European Conference, {ECML} {PKDD} 2025, Porto, Portugal, September
                  15-19, 2025, Proceedings, Part {V}},
  series       = {Lecture Notes in Computer Science},
  volume       = {16017},
  pages        = {276--294},
  publisher    = {Springer},
  year         = {2025},
  url          = {https://doi.org/10.1007/978-3-032-06096-9\_16},
  doi          = {10.1007/978-3-032-06096-9\_16},
  bibsource    = {dblp computer science bibliography, https://dblp.org}
}

@inproceedings{DBLP:conf/aaai/FosterSB24,
  author       = {Jack Foster and
                  Stefan Schoepf and
                  Alexandra Brintrup},
  editor       = {Michael J. Wooldridge and
                  Jennifer G. Dy and
                  Sriraam Natarajan},
  title        = {Fast Machine Unlearning without Retraining through Selective Synaptic
                  Dampening},
  booktitle    = {Thirty-Eighth {AAAI} Conference on Artificial Intelligence, {AAAI}
                  2024, Thirty-Sixth Conference on Innovative Applications of Artificial
                  Intelligence, {IAAI} 2024, Fourteenth Symposium on Educational Advances
                  in Artificial Intelligence, {EAAI} 2024, February 20-27, 2024, Vancouver,
                  Canada},
  pages        = {12043--12051},
  publisher    = {{AAAI} Press},
  year         = {2024},
  url          = {https://doi.org/10.1609/aaai.v38i11.29092},
  doi          = {10.1609/AAAI.V38I11.29092},
  bibsource    = {dblp computer science bibliography, https://dblp.org}
}

@inproceedings{DBLP:conf/iclr/FosterSB24,
  author       = {Jack Foster and
                  Stefan Schoepf and
                  Alexandra Brintrup},
  title        = {Loss-Free Machine Unlearning},
  booktitle    = {The Second Tiny Papers Track at {ICLR} 2024, Tiny Papers @ {ICLR}
                  2024, Vienna, Austria, May 11, 2024},
  publisher    = {OpenReview.net},
  year         = {2024},
  url          = {https://openreview.net/forum?id=bCPz7uqmmh},
  bibsource    = {dblp computer science bibliography, https://dblp.org}
}

@article{DBLP:journals/tnn/TarunCMK24,
  author       = {Ayush K. Tarun and
                  Vikram S. Chundawat and
                  Murari Mandal and
                  Mohan S. Kankanhalli},
  title        = {Fast Yet Effective Machine Unlearning},
  journal      = {{IEEE} Trans. Neural Networks Learn. Syst.},
  volume       = {35},
  number       = {9},
  pages        = {13046--13055},
  year         = {2024},
  url          = {https://doi.org/10.1109/TNNLS.2023.3266233},
  doi          = {10.1109/TNNLS.2023.3266233},
  bibsource    = {dblp computer science bibliography, https://dblp.org}
}

@inproceedings{DBLP:conf/icb/ShivamNRG25,
  author       = {FNU Shivam and
                  Nima Najafzadeh and
                  Yenumula Reddy and
                  Prashnna K. Gyawali},
  title        = {{CURE:} Centroid-guided Unsupervised Representation Erasure for Facial
                  Recognition Systems},
  booktitle    = {{IEEE} International Joint Conference on Biometrics, {IJCB} 2025,
                  Osaka, Japan, September 8-11, 2025},
  pages        = {1--10},
  publisher    = {{IEEE}},
  year         = {2025},
  url          = {https://doi.org/10.1109/IJCB65343.2025.11410631},
  doi          = {10.1109/IJCB65343.2025.11410631},
  bibsource    = {dblp computer science bibliography, https://dblp.org}
}

@inproceedings{DBLP:conf/cikm/LeeW23,
  author       = {Sangyong Lee and
                  Simon S. Woo},
  editor       = {Ingo Frommholz and
                  Frank Hopfgartner and
                  Mark Lee and
                  Michael Oakes and
                  Mounia Lalmas and
                  Min Zhang and
                  Rodrygo L. T. Santos},
  title        = {{UNDO:} Effective and Accurate Unlearning Method for Deep Neural Networks},
  booktitle    = {Proceedings of the 32nd {ACM} International Conference on Information
                  and Knowledge Management, {CIKM} 2023, Birmingham, United Kingdom,
                  October 21-25, 2023},
  pages        = {4043--4047},
  publisher    = {{ACM}},
  year         = {2023},
  url          = {https://doi.org/10.1145/3583780.3615235},
  doi          = {10.1145/3583780.3615235},
  bibsource    = {dblp computer science bibliography, https://dblp.org}
}

@INPROCEEDINGS{10426386Compromising,
  author={Nair, Adithyan M and Kumar, Akshit Sudheer and Kumar, Devakrishna Sanil and T, Anjali},
  booktitle={2023 3rd International Conference on Innovative Mechanisms for Industry Applications (ICIMIA)}, 
  title={Selective Unlearning in Face Recognition: Forgetting Faces without Compromising Accuracy}, 
  year={2023},
  volume={},
  number={},
  pages={211-216},
  doi={10.1109/ICIMIA60377.2023.10426386}}

@inproceedings{DBLP:conf/wacv/Zakharov26,
  author       = {Mikhail Zakharov},
  title        = {Face Identity Unlearning for Retrieval via Embedding Dispersion},
  booktitle    = {{IEEE/CVF} Winter Conference on Applications of Computer Vision, {WACV}
                  2026 - Workshops, Tucson, AZ, USA, March 6-10, 2026},
  pages        = {1277--1286},
  publisher    = {{IEEE}},
  year         = {2026},
  url          = {https://doi.org/10.1109/WACVW68408.2026.00139},
  doi          = {10.1109/WACVW68408.2026.00139},
  bibsource    = {dblp computer science bibliography, https://dblp.org}
}

@misc{ozturk2026damnatiomemoriaeadversariallyselectively,
      title={Damnatio Memoriae: Adversarially and Selectively Forgetting Identities in the Embedding Space of Face Recognition Models}, 
      author={Ünsal Öztürk and Vedrana Krivokuća Hahn and Sushil Bhattacharjee and Sébastien Marcel},
      year={2026},
      eprint={2609.27115},
      archivePrefix={arXiv},
      primaryClass={cs.CV},
      url={https://arxiv.org/abs/2609.27115}, 
}

@InProceedings{MS_celeb_1m,
    author="Guo, Yandong
    and Zhang, Lei
    and Hu, Yuxiao
    and He, Xiaodong
    and Gao, Jianfeng",
    editor="Leibe, Bastian
    and Matas, Jiri
    and Sebe, Nicu
    and Welling, Max",
    title="MS-Celeb-1M: A Dataset and Benchmark for Large-Scale Face Recognition",
    booktitle="Computer Vision -- ECCV 2016",
    year="2016",
    publisher="Springer International Publishing",
    address="Cham",
    pages="87--102",
    isbn="978-3-319-46487-9"
}

@inproceedings{huang_inria-00321923,
  TITLE = {{Labeled Faces in the Wild: A Database for Studying Face Recognition in Unconstrained Environments}},
  AUTHOR = {Huang, Gary B. and Mattar, Marwan and Berg, Tamara and Learned-Miller, Eric},
  URL = {https://inria.hal.science/inria-00321923},
  BOOKTITLE = {{Workshop on Faces in 'Real-Life' Images: Detection, Alignment, and Recognition}},
  ADDRESS = {Marseille, France},
  ORGANIZATION = {{Erik Learned-Miller and Andras Ferencz and Fr{\'e}d{\'e}ric Jurie}},
  YEAR = {2008},
  MONTH = Oct,
  HAL_ID = {inria-00321923},
  HAL_VERSION = {v1},
}

@INPROCEEDINGS{cfpfp2016Sengupta,
  author={Sengupta, Soumyadip and Chen, Jun-Cheng and Castillo, Carlos and Patel, Vishal M. and Chellappa, Rama and Jacobs, David W.},
  booktitle={2016 IEEE Winter Conference on Applications of Computer Vision (WACV)}, 
  title={Frontal to profile face verification in the wild}, 
  year={2016},
  volume={},
  number={},
  pages={1-9},
  doi={10.1109/WACV.2016.7477558}
}

@inproceedings{moschoglou2017agedb,
  title={Agedb: the first manually collected, in-the-wild age database},
  author={Moschoglou, Stylianos and Papaioannou, Athanasios and Sagonas, Christos and Deng, Jiankang and Kotsia, Irene and Zafeiriou, Stefanos},
  booktitle={proceedings of the IEEE conference on computer vision and pattern recognition workshops},
  pages={51--59},
  year={2017}
}

@TechReport{CPLFWTech,
  author =       {T. Zheng and W. Deng},
  title =        {Cross-pose LFW: A database for studying cross-pose face recognition in unconstrained environments},
  institution =  {Beijing University of Posts and Telecommunications},
  year =         {2018},
  number =       {18-01},
  month =        {February}
}

@article{abs-1708-08197,
  author    = {Tianyue Zheng and
               Weihong Deng and
               Jiani Hu},
  title     = {Cross-Age {LFW:} {A} Database for Studying Cross-Age Face Recognition
               in Unconstrained Environments},
  journal   = {CoRR},
  volume    = {abs/1708.08197},
  year      = {2017},
  eprint    = {1708.08197},
}

@inproceedings{WhitelamTBMAMKJ17,
  author       = {Cameron Whitelam and
                  Emma Taborsky and
                  Austin Blanton and
                  Brianna Maze and
                  Jocelyn C. Adams and
                  Tim Miller and
                  Nathan D. Kalka and
                  Anil K. Jain and
                  James A. Duncan and
                  Kristen Allen and
                  Jordan Cheney and
                  Patrick Grother},
  title        = {{IARPA} Janus Benchmark-B Face Dataset},
  booktitle    = {2017 {IEEE} Conference on Computer Vision and Pattern Recognition
                  Workshops, {CVPR} Workshops 2017, Honolulu, HI, USA, July 21-26, 2017},
  pages        = {592--600},
  publisher    = {{IEEE} Computer Society},
  year         = {2017},
  url          = {https://doi.org/10.1109/CVPRW.2017.87},
  doi          = {10.1109/CVPRW.2017.87},
}

@inproceedings{MazeADKMO0NACG18,
  author       = {Brianna Maze and
                  Jocelyn C. Adams and
                  James A. Duncan and
                  Nathan D. Kalka and
                  Tim Miller and
                  Charles Otto and
                  Anil K. Jain and
                  W. Tyler Niggel and
                  Janet Anderson and
                  Jordan Cheney and
                  Patrick Grother},
  title        = {{IARPA} Janus Benchmark - {C:} Face Dataset and Protocol},
  booktitle    = {2018 International Conference on Biometrics, {ICB} 2018, Gold Coast,
                  Australia, February 20-23, 2018},
  pages        = {158--165},
  publisher    = {{IEEE}},
  year         = {2018},
  url          = {https://doi.org/10.1109/ICB2018.2018.00033},
  doi          = {10.1109/ICB2018.2018.00033},
}

@inproceedings{deng2019arcface,
  title={Arcface: Additive angular margin loss for deep face recognition},
  author={Deng, Jiankang and Guo, Jia and Xue, Niannan and Zafeiriou, Stefanos},
  booktitle={Proceedings of the IEEE/CVF conference on computer vision and pattern recognition},
  pages={4690--4699},
  year={2019}
}

@INPROCEEDINGS{9880230,
  author={Kim, Minchul and Jain, Anil K. and Liu, Xiaoming},
  booktitle={2022 IEEE/CVF Conference on Computer Vision and Pattern Recognition (CVPR)}, 
  title={AdaFace: Quality Adaptive Margin for Face Recognition}, 
  year={2022},
  volume={},
  number={},
  pages={18729-18738},
  doi={10.1109/CVPR52688.2022.01819}
}

@inproceedings{meng2021magface,
  title     = {MagFace: A Universal Representation for Face Recognition and Quality Assessment},
  author    = {Meng, Qingliang and Zhao, Shichao and Huang, Zhida and Zhou, Feng},
  booktitle = {Proceedings of the IEEE/CVF Conference on Computer Vision and Pattern Recognition (CVPR)},
  pages     = {14225--14234},
  year      = {2021}
}

@INPROCEEDINGS{10030931,
  author={Terhörst, Philipp and Ihlefeld, Malte and Huber, Marco and Damer, Naser and Kirchbuchner, Florian and Raja, Kiran and Kuijper, Arjan},
  booktitle={2023 IEEE/CVF Winter Conference on Applications of Computer Vision (WACV)}, 
  title={QMagFace: Simple and Accurate Quality-Aware Face Recognition}, 
  year={2023},
  volume={},
  number={},
  pages={3473-3483},
  doi={10.1109/WACV56688.2023.00348}
}

@inproceedings{DBLP:conf/eccv/BelabbaciWT26,
  author       = {El Ouanas Belabbaci and
                  Bhavesh Wani and
                  Philipp Terh{\"{o}}rst},
  editor       = {Paolo Favaro and
                  Zuzana Kukelova and
                  Atsuto Maki and
                  Anna Rohrbach and
                  Konrad Schindler and
                  Federico Tombari},
  title        = {Learning to Attract and Repel: Dual Quality Margin Learning for Face
                  Recognition (DQM-Face)},
  booktitle    = {Computer Vision - {ECCV} 2026 - 19th European Conference, Malm{\"{o}},
                  Sweden, September 8-12, 2026, Proceedings, Part {XXVI}},
  series       = {Lecture Notes in Computer Science},
  volume       = {17026},
  pages        = {40--58},
  publisher    = {Springer},
  year         = {2026},
  url          = {https://doi.org/10.1007/978-3-032-37595-7\_3},
  doi          = {10.1007/978-3-032-37595-7\_3},
  bibsource    = {dblp computer science bibliography, https://dblp.org}
}

@standard{ISO_IEC_19795_1_2006,
  author       = {{International Organization for Standardization}},
  title        = {{ISO/IEC 19795-1:2006 Information technology -- Biometric performance testing and reporting -- Part 1: Principles and framework}},
  year         = {2006},
  institution  = {{International Organization for Standardization}},
  type         = {International Standard},
  number       = {ISO/IEC 19795-1:2006}
}

@InProceedings{Schroff_2015_CVPR,
author = {Schroff, Florian and Kalenichenko, Dmitry and Philbin, James},
title = {FaceNet: A Unified Embedding for Face Recognition and Clustering},
booktitle = {Proceedings of the IEEE Conference on Computer Vision and Pattern Recognition (CVPR)},
month = {June},
year = {2015}
}

@inproceedings{DBLP:conf/icb/AlRefaiRRCT25,
  author       = {Rouqaiah Al{-}Refai and
                  Pankaja Priya Ramasamy and
                  Ragini Ramesh and
                  Patricia Arias Cabarcos and
                  Philipp Terh{\"{o}}rst},
  title        = {A Comprehensive Re-Evaluation of Biometric Modality Properties in
                  the Modern Era},
  booktitle    = {{IEEE} International Joint Conference on Biometrics, {IJCB} 2025,
                  Osaka, Japan, September 8-11, 2025},
  pages        = {1--11},
  publisher    = {{IEEE}},
  year         = {2025},
  url          = {https://doi.org/10.1109/IJCB65343.2025.11411391},
  doi          = {10.1109/IJCB65343.2025.11411391},
  bibsource    = {dblp computer science bibliography, https://dblp.org}
}

@article{10.1145/3736753,
author = {Leyva, Roberto and Gregory, Epiphaniou and Maple, Carsten},
title = {Attack Vectors for Face Recognition Systems: A Comprehensive Review},
year = {2025},
issue_date = {January 2026},
publisher = {Association for Computing Machinery},
address = {New York, NY, USA},
volume = {58},
number = {1},
issn = {0360-0300},
url = {https://doi.org/10.1145/3736753},
doi = {10.1145/3736753},
journal = {ACM Comput. Surv.},
month = sep,
articleno = {19},
numpages = {37}
}

@inproceedings{DBLP:conf/icb/TerhorstFDKK20,
  author       = {Philipp Terh{\"{o}}rst and
                  Daniel F{\"{a}}hrmann and
                  Naser Damer and
                  Florian Kirchbuchner and
                  Arjan Kuijper},
  title        = {Beyond Identity: What Information Is Stored in Biometric Face Templates?},
  booktitle    = {2020 {IEEE} International Joint Conference on Biometrics, {IJCB} 2020,
                  Houston, TX, USA, September 28 - October 1, 2020},
  pages        = {1--10},
  publisher    = {{IEEE}},
  year         = {2020},
  url          = {https://doi.org/10.1109/IJCB48548.2020.9304874},
  doi          = {10.1109/IJCB48548.2020.9304874},
  bibsource    = {dblp computer science bibliography, https://dblp.org}
}

@article{DBLP:journals/ivc/VaretoLBSG24,
  author       = {Rafael Henrique Vareto and
                  Yu Linghu and
                  Terrance E. Boult and
                  William Robson Schwartz and
                  Manuel G{\"{u}}nther},
  title        = {Open-set face recognition with maximal entropy and Objectosphere loss},
  journal      = {Image Vis. Comput.},
  volume       = {141},
  pages        = {104862},
  year         = {2024},
  url          = {https://doi.org/10.1016/j.imavis.2023.104862},
  doi          = {10.1016/J.IMAVIS.2023.104862},
  bibsource    = {dblp computer science bibliography, https://dblp.org}
}
}

\clearpage
\setcounter{page}{1}
\maketitlesupplementary
\appendix

This supplementary material provides additional methodological details, experimental supporting results and discussions for the proposed unlearning framework and its evaluation. We first present the optimization objective underlying the closed-form regularized identity suppression (RIS) strategy, then give per-model training details for the linearizer framework. We further extend the evaluation to additional iResNet50 backbones to test generalization across backbone capacities. Finally, we discuss the limitations of closed-set unlearning metrics for evaluating identity-level unlearning in open-set FR. 

\section{Regularized Identity Suppression Objective}
The RIS unlearning strategy described in Section \ref{sec:method_ris} is based on a closed-form solution derived from a regularized weighted least-squares problem. The underlying optimization formulation is presented here to clarify the mathematical basis of this solution and complement the method description with further details on its derivation in the context of identity unlearning.

The main objectives of RIS optimization are to minimize the deviation of forget representations from $z^\star$ while preserving the original mappings of retain representations. The optimization problem is defined as:

\begin{equation}
\begin{aligned}
A' = \arg\min_{\hat{A}} \Bigg[
&\frac{1}{n_f}\sum_{i=1}^{n_f}
\left\|\hat{A}\hat{e}_i-z^\star\right\|_2^2 \\
&+\frac{\lambda}{n_r}\sum_{j=1}^{n_r}
\left\|\hat{A}\hat{e}_j-A\hat{e}_j\right\|_2^2
+\varepsilon\|\hat{A}\|_F^2
\Bigg].
\end{aligned}
\label{eq:ris_objective}
\end{equation}

In which $\hat{A}$ denotes the candidate linear mapping.  Differentiating the objective with respect to $\hat{A}$ and setting the gradient to zero yields:
\begin{equation}
\begin{aligned}
\hat{A}\Bigg(
&\frac{1}{n_f}\sum_{i=1}^{n_f}
\hat{e}_i\hat{e}_i^\top
+\frac{\lambda}{n_r}\sum_{j=1}^{n_r}
\hat{e}_j\hat{e}_j^\top
+\varepsilon I
\Bigg) \\
&=z^\star\mu_f^\top
+\frac{\lambda}{n_r}\sum_{j=1}^{n_r}
A\hat{e}_j\hat{e}_j^\top.
\end{aligned}
\label{eq:ris_normal}
\end{equation}

Using the definitions of $B$ and $G$ from the main paper, this simplifies to $\hat{A}G=B$. Solving for the optimal mapping leads to the closed-form solution $A'=BG^{-1}$.

\section{Linearizer Training Fidelity}
We further complement the linearizer validation described in Section \ref{sec:implementation} by reporting the detailed reconstruction cosine similarity and inversion error for each evaluated FR model. As shown in Table \ref{tab:linearizer_fidelity}  all models achieve reconstruction cosine similarities above $0.99$ and inversion errors below $10^{-6}$, satisfying the validation criteria defined in the main paper and confirming that the linearizer produces nearly identical face templates to the original model. 

\begin{table}[h!]
\centering
\footnotesize
\caption{\textbf{Linearizer reconstruction fidelity and invertibility.}
Reconstruction cosine similarity (Recon. cos) and inversion error (Inv. err) across the evaluated FR models.}
\label{tab:linearizer_fidelity}
\begin{tabular}{lcc}
\toprule
Model & Recon. cos & Inv. err \\
\midrule
ArcFace100   & 0.999972 & $2.5 \times 10^{-7}$ \\
ArcFace50    & 0.999999 & $1.49 \times 10^{-7}$ \\
AdaFace100   & 0.999996 & $2.1 \times 10^{-7}$ \\
AdaFace50    & 0.999951 & $2.8 \times 10^{-7}$ \\
MagFace100   & 0.999964 & $2.7 \times 10^{-7}$ \\
MagFace50    & 0.999927 & $3.6 \times 10^{-7}$ \\
DQMFace100  & 0.999998 & $8.94 \times 10^{-8}$ \\
\bottomrule
\end{tabular}
\end{table}

\section{Extended Experimental Results}
To further assess the generalizability of the linearizer framework and unlearning strategies across different backbone capacities, we extend the experiments to smaller backbone (iResNet50) models, complementing the iResNet100 results discussed in Section \ref{sec:results}. Since the iResNet50 results show similar trends to those observed for iResNet100, their detailed numerical evaluation is provided in this section to avoid repeating similar findings. In general, both IS$^3$ and RIS show consistent behavior with iResNet100 results, maintaining high unlearning effectiveness at small forget identity scales $|\mathcal{C}_{f}|=\{10, 50\}$, which gradually declines when the number of unlearned identities increases. Similar differences between the methods are observed with iResNet50 models. For instance, RIS is better at preserving face verification performance at larger forget scales for models such as AdaFace, MagFace, and QMagFace. In contrast, IS$^3$ is more sensitive to the number of unlearned identities, showing greater degradation in both verification and unlearning performance when 10-20\% of the training identities are unlearned. These observations are supported by the results in Tables \ref{tab:unlearning_r50}, \ref{tab:verification_benchmarks_r50} and \ref{tab:ijb_r50}, which report the linearizer recognition preservation, unlearning effectiveness under closed-set and open-set evaluation, as well as face verification performance across standard benchmarks.

\begin{table*}[t]
\centering
\footnotesize
\setlength{\tabcolsep}{4pt}
\renewcommand{\arraystretch}{0.4}
\caption{\textbf{Unlearning performance evaluation}.
Results are reported for iResNet50 FR models across varying numbers of forgotten identities, considering both closed-set and open-set unlearning efficiency, and change in recognition performance. ($\Delta\mathrm{Avg}_5$) reports the change in mean verification accuracy across five benchmarks as percentage points relative to the corresponding Lin and Base performance.}
\label{tab:unlearning_r50}
\begin{tabular}{@{}cl*{10}{c}@{}}
\toprule
& & \multicolumn{5}{c}{IS$^3$} & \multicolumn{5}{c}{RIS} \\
\cmidrule(lr){3-7}\cmidrule(l){8-12} & & \multicolumn{2}{c}{Closed-set} & \multicolumn{2}{c}{Open-set}
    & & \multicolumn{2}{c}{Closed-set} & \multicolumn{2}{c}{Open-set} & \\
\cmidrule(lr){3-4}\cmidrule(lr){5-6} \cmidrule(lr){8-9}\cmidrule(lr){10-11}
Model & $|\mathcal{C}_{f}|$
& Fgt. \% & Ret. \% & FS & $\Delta \text{Avg}_5$ & Time (s)
& Fgt. \% & Ret. \% & FS & $\Delta \text{Avg}_5$ & Time (s) \\
\midrule
\multirow{12}{*}{\rotatebox[origin=c]{90}{\strut ArcFace}}
& \textit{Base} & 100.00 & 98.94 & 0.000 & - & -
       & 100.00 & 98.94 & 0.000 & - & - \\
& \textit{Lin} & 100.00 & 98.93 & 0.000 & - & -
       & 100.00 & 98.93 & 0.000 & - & - \\
& 1    & 100.00 & 98.93 & 0.989 & +0.04 & 0.23
       & 0.00 & 98.93 & 1.000 & +0.04 & 0.28 \\
& 5    & 20.22 & 98.93 & 1.000 & -0.03 & 0.19
       & 0.00 & 98.92 & 1.000 & -0.08 & 0.28 \\
& 10   & 1.06 & 98.93 & 1.000 & -0.16 & 0.19
       & 0.00 & 98.90 & 1.000 & -0.29 & 0.27 \\
& 50   & 0.75 & 98.88 & 0.997 & -0.14 & 0.25
       & 0.00 & 98.76 & 1.000 & -2.36 & 0.33 \\
& 100  & 0.31 & 98.75 & 0.997 & -0.57 & 0.26
       & 0.10 & 98.10 & 0.997 & -12.93 & 0.28 \\
& 500  & 72.79 & 98.44 & 0.886 & -1.25 & 0.28
       & 3.83 & 76.63 & 0.755 & -19.62 & 0.29 \\
& 1K & 96.73 & 98.56 & 0.682 & -1.26 & 0.31
       & 7.51 & 57.67 & 0.555 & -20.60 & 0.31 \\
\midrule
\multirow{12}{*}{\rotatebox[origin=c]{90}{\strut AdaFace}}
& \textit{Base} & 100.00 & 98.95 & 0.000 & - & -
       & 100.00 & 98.95 & 0.000 & - & - \\
& \textit{Lin} & 100.00 & 98.98 & 0.000 & - & -
       & 100.00 & 98.98 & 0.000 & - & - \\
& 1    & 100.00 & 98.98 & 0.728 & -0.04 & 0.20
       & 0.00 & 98.97 & 1.000 & -0.07 & 0.29 \\
& 5    & 24.72 & 98.97 & 0.993 & -0.15 & 0.22
       & 0.00 & 98.97 & 1.000 & -0.21 & 0.30 \\
& 10   & 2.65 & 98.96 & 0.994 & -0.25 & 0.22
       & 0.53 & 98.96 & 1.000 & -0.37 & 0.31 \\
& 50   & 10.20 & 98.96 & 0.973 & -0.28 & 0.23
       & 1.50 & 98.86 & 0.996 & -1.24 & 0.31 \\
& 100  & 2.62 & 98.61 & 0.964 & -1.00 & 0.24
       & 7.17 & 98.78 & 0.994 & -1.31 & 0.30 \\
& 500  & 18.34 & 83.77 & 0.516 & -3.14 & 0.27
       & 98.09 & 98.77 & 0.933 & -1.12 & 0.32 \\
& 1K & 46.72 & 81.54 & 0.301 & -3.36 & 0.32
       & 98.62 & 98.91 & 0.839 & -0.94 & 0.32 \\
\midrule
\multirow{12}{*}{\rotatebox[origin=c]{90}{\strut MagFace}}
& \textit{Base} & 100.00 & 98.93 & 0.000 & - & -
       & 100.00 & 98.93 & 0.000 & - & - \\
& \textit{Lin} & 100.00 & 98.94 & 0.000 & - & -
       & 100.00 & 98.94 & 0.000 & - & - \\
& 1    & 100.00 & 98.93 & 0.804 & -0.03 & 0.18
       & 0.00 & 98.94 & 1.000 & -0.03 & 0.29 \\
& 5    & 1.12 & 98.92 & 1.000 & -0.05 & 0.19
       & 0.00 & 98.90 & 1.000 & -0.24 & 0.32 \\
& 10   & 0.00 & 98.90 & 1.000 & -0.17 & 0.18
       & 0.00 & 98.89 & 1.000 & -0.32 & 0.28 \\
& 50   & 0.11 & 98.82 & 0.994 & -0.24 & 0.20
       & 0.00 & 98.70 & 1.000 & -1.05 & 0.33 \\
& 100  & 0.00 & 96.35 & 0.981 & -2.72 & 0.21
       & 0.05 & 98.00 & 0.988 & -1.73 & 0.33 \\
& 500  & 0.21 & 0.06 & 0.013 & -4.89 & 0.22
       & 33.55 & 88.39 & 0.604 & -1.45 & 0.29 \\
& 1K & 0.10 & 0.06 & 0.035 & -7.46 & 0.24
       & 40.85 & 78.90 & 0.403 & -1.10 & 0.30 \\
\midrule
\multirow{12}{*}{\rotatebox[origin=c]{90}{\strut QMagFace}}
& \textit{Base} & 100.00 & 98.93 & 0.000 & - & -
       & 100.00 & 98.93 & 0.000 & - & - \\
& \textit{Lin} & 100.00 & 98.94 & 0.000 & - & -
       & 100.00 & 98.94 & 0.000 & - & - \\
& 1    & 100.00 & 98.93 & 0.804 & -0.07 & 0.18
       & 0.00 & 98.94 & 1.000 & -0.06 & 0.29 \\
& 5    & 1.12 & 98.92 & 1.000 & -0.02 & 0.19
       & 0.00 & 98.90 & 1.000 & -0.15 & 0.32 \\
& 10   & 0.00 & 98.90 & 1.000 & -0.20 & 0.18
       & 0.00 & 98.89 & 1.000 & -0.17 & 0.28 \\
& 50   & 0.11 & 98.82 & 0.994 & -0.30 & 0.20
       & 0.00 & 98.70 & 1.000 & -1.19 & 0.33 \\
& 100  & 0.00 & 96.35 & 0.981 & -2.62 & 0.21
       & 0.05 & 98.00 & 0.988 & -2.01 & 0.33 \\
& 500  & 0.21 & 0.06 & 0.013 & -2.35 & 0.22
       & 33.55 & 88.39 & 0.604 & -1.71 & 0.29 \\
& 1K & 0.10 & 0.06 & 0.035 & -2.37 & 0.24
       & 40.85 & 78.90 & 0.403 & -1.34 & 0.30 \\
\bottomrule
\end{tabular}
\end{table*}

\begin{table*}[t]
\footnotesize
\caption{\textbf{Impact of unlearning on small-scale face verification performance.} 
Results are reported for iResNet50 FR models across five verification benchmarks, showing changes in mean accuracy after unlearning relative to the corresponding linearized model (Lin) performance. Changes are expressed in percentage points, with negative values indicating performance degradation.}
\label{tab:verification_benchmarks_r50}
\centering
\setlength{\tabcolsep}{4pt}
\renewcommand{\arraystretch}{0.4}
\begin{tabular}{@{}clcccccccccc@{}}
\toprule
& & \multicolumn{5}{c}{IS$^3$} & \multicolumn{5}{c}{RIS} \\
\cmidrule(lr){3-7} \cmidrule(l){8-12}
Model & $|\mathcal{C}_{f}|$ & LFW & CFP-FP & AgeDB & CALFW & CPLFW & LFW & CFP-FP & AgeDB & CALFW & CPLFW \\
\midrule
\multirow{14}{*}{\rotatebox{90}{ArcFace}} 
& \textit{Base} & 99.78 & 97.03 & 97.70 & 95.83 & 91.80 & 99.78 & 97.03 & 97.70 & 95.83 & 91.80 \\
& \textit{Lin} & 99.78 & 97.04 & 97.70 & 95.83 & 91.82 & 99.78 & 97.04 & 97.70 & 95.83 & 91.82 \\
& 1 & +0.00 & +0.10 & +0.02 & +0.10 & -0.02 & +0.00 & +0.00 & +0.05 & +0.15 & -0.10 \\
& 5 & +0.00 & -0.06 & +0.20 & -0.02 & -0.30 & -0.02 & -0.10 & -0.05 & -0.05 & -0.20 \\
& 10 & -0.05 & -0.27 & +0.03 & -0.03 & -0.50 & -0.08 & -0.77 & -0.28 & +0.08 & -0.38 \\
& 50 & -0.02 & -0.14 & +0.13 & -0.12 & -0.53 & -0.23 & -4.43 & -2.62 & -1.43 & -3.10 \\
& 100 & -0.07 & -0.93 & -0.55 & -0.28 & -1.00 & -5.83 & -18.33 & -16.40 & -10.13 & -13.97 \\
& 500 & -0.20 & -2.44 & -1.13 & -0.73 & -1.73 & -11.92 & -25.20 & -22.47 & -17.77 & -20.75 \\
& 1K & -0.12 & -2.19 & -1.53 & -0.78 & -1.68 & -12.87 & -26.13 & -23.42 & -18.48 & -22.12 \\
\midrule
\multirow{14}{*}{\rotatebox{90}{AdaFace}} 
& \textit{Base} & 99.78 & 97.77 & 98.07 & 96.13 & 92.63 & 99.78 & 97.77 & 98.07 & 96.13 & 92.63 \\
& \textit{Lin} & 99.78 & 97.89 & 98.08 & 96.13 & 92.55 & 99.78 & 97.89 & 98.08 & 96.13 & 92.55 \\
& 1 & +0.00 & -0.07 & -0.03 & -0.08 & -0.03 & +0.00 & -0.23 & -0.12 & -0.07 & +0.07 \\
& 5 & -0.03 & -0.14 & -0.30 & -0.08 & -0.20 & -0.02 & -0.34 & -0.32 & -0.10 & -0.28 \\
& 10 & -0.12 & -0.46 & -0.23 & -0.25 & -0.20 & -0.02 & -0.60 & -0.72 & -0.18 & -0.35 \\
& 50 & -0.02 & -0.74 & -0.23 & -0.15 & -0.23 & -0.10 & -2.23 & -1.50 & -0.62 & -1.77 \\
& 100 & -0.18 & -1.43 & -1.58 & -0.48 & -1.33 & -0.15 & -2.14 & -1.73 & -0.75 & -1.77 \\
& 500 & -0.37 & -4.70 & -5.02 & -2.22 & -3.40 & -0.22 & -1.93 & -1.23 & -0.60 & -1.63 \\
& 1K & -0.52 & -4.90 & -5.40 & -2.68 & -3.28 & -0.17 & -1.67 & -1.23 & -0.40 & -1.25 \\
\midrule
\multirow{14}{*}{\rotatebox{90}{MagFace}} 
& \textit{Base} & 99.82 & 97.44 & 97.88 & 96.07 & 92.00 &  99.82 & 97.44 & 97.88 & 96.07 & 92.00 \\
& \textit{Lin} & 99.80 & 97.93 & 97.92 & 96.02 & 92.43 & 99.80 & 97.93 & 97.92 & 96.02 & 92.43 \\
& 1 & +0.00 & -0.11 & -0.07 & -0.03 & +0.07 & +0.00 & -0.11 & -0.08 & -0.03 & -0.07 \\
& 5 & +0.00 & -0.20 & -0.05 & +0.05 & -0.03 & -0.02 & -0.39 & -0.38 & -0.18 & -0.22 \\
& 10 & -0.02 & -0.14 & -0.35 & -0.02 & -0.30 & +0.00 & -0.33 & -0.72 & -0.27 & -0.28 \\
& 50 & -0.02 & -0.30 & -0.33 & -0.12 & -0.43 & -0.13 & -1.80 & -1.27 & -0.52 & -1.55 \\
& 100 & -0.52 & -3.79 & -3.78 & -2.28 & -3.25 & -0.18 & -2.54 & -2.35 & -1.07 & -2.50 \\
& 500 & -4.28 & -4.76 & -6.72 & -3.95 & -4.73 & -0.25 & -2.29 & -1.85 & -0.60 & -2.27 \\
& 1K & -7.75 & -6.67 & -10.28 & -6.88 & -5.73 & -0.12 & -1.89 & -1.32 & -0.40 & -1.80 \\
\midrule
\multirow{14}{*}{\rotatebox{90}{QMagFace}} 
& \textit{Base} & 99.80 & 98.03 & 98.02 & 96.05 & 92.98 & 99.80 & 98.03 & 98.02 & 96.05 & 92.98 \\
& \textit{Lin} & 99.78 & 98.26 & 98.03 & 96.15 & 93.13 & 99.78 & 98.26 & 98.03 & 96.15 & 93.13 \\
& 1 & +0.02 & -0.09 & -0.05 & -0.05 & -0.20 & +0.02 & +0.01 & +0.12 & -0.02 & -0.22 \\
& 5 & -0.02 & -0.01 & -0.10 & +0.03 & +0.00 & +0.00 & -0.21 & -0.20 & -0.15 & -0.20 \\
& 10 & +0.02 & -0.10 & -0.10 & -0.23 & -0.60 & +0.03 & -0.10 & -0.45 & -0.10 & -0.23 \\
& 50 & -0.08 & -0.14 & -0.57 & -0.25 & -0.47 & -0.10 & -1.70 & -1.30 & -0.67 & -2.18 \\
& 100 & -0.30 & -3.20 & -3.88 & -2.15 & -3.55 & -0.20 & -2.76 & -2.50 & -1.20 & -3.43 \\
& 500 & -0.17 & -3.83 & -2.70 & -1.08 & -3.98 & -0.23 & -2.60 & -1.85 & -0.77 & -3.10 \\
& 1K & -0.12 & -4.01 & -2.53 & -1.18 & -4.00 & -0.15 & -2.21 & -1.35 & -0.42 & -2.55 \\
\bottomrule
\end{tabular}
\end{table*}

\begin{table*}[t]
\footnotesize
\caption{\textbf{Impact of unlearning on large-scale face verification performance.}
Results are reported for iResNet50 FR models on IJB-B and IJB-C using TAR at FAR thresholds of $10^{-5}$, $10^{-4}$, and $10^{-3}$. The Base and Lin performance is reported in absolute values, while changes after unlearning are expressed in percentage points relative to the corresponding Lin, with negative values indicating degradation.}
\label{tab:ijb_r50}
\centering
\setlength{\tabcolsep}{4pt}
\renewcommand{\arraystretch}{0.4}
\begin{tabular}{@{}clcccccccccccc@{}}
\toprule
& & \multicolumn{6}{c}{IS$^3$} & \multicolumn{6}{c}{RIS} \\
\cmidrule(lr){3-8}\cmidrule(l){9-14}
& & \multicolumn{3}{c}{IJB-B} & \multicolumn{3}{c}{IJB-C} & \multicolumn{3}{c}{IJB-B} & \multicolumn{3}{c}{IJB-C} \\
\cmidrule(lr){3-5}\cmidrule(lr){6-8}\cmidrule(lr){9-11}\cmidrule(l){12-14}
Model & $|\mathcal{C}_{f}|$ & $10^{-5}$ & $10^{-4}$ & $10^{-3}$ & $10^{-5}$ & $10^{-4}$ & $10^{-3}$ & $10^{-5}$ & $10^{-4}$ & $10^{-3}$ & $10^{-5}$ & $10^{-4}$ & $10^{-3}$ \\
\midrule
\multirow{14}{*}{\rotatebox{90}{ArcFace}} 
& \textit{Base} & 87.12 & 94.03 & 96.22 & 92.60 & 95.67 & 97.20 & 87.12 & 94.03 & 96.22 & 92.60 & 95.67 & 97.20 \\
& \textit{Lin} & 86.73 & 93.87 & 96.21 & 91.83 & 95.50 & 97.16 & 86.73 & 93.87 & 96.21 & 91.83 & 95.50 & 97.16 \\
& 1 & -0.33 & -0.07 & +0.02 & -0.01 & -0.02 & +0.01 
& -0.07 & -0.16 & +0.01 & -0.03 & -0.04 & -0.04 \\
& 5 & +0.18 & -0.06 & +0.02 & +0.10 & -0.03 & -0.01 
& -0.36 & -0.32 & -0.22 & -0.26 & -0.20 & -0.20 \\
& 10 & +0.10 & -0.11 & -0.09 & -0.01 & -0.08 & -0.11 
& -0.89 & -0.64 & -0.47 & -0.66 & -0.54 & -0.36 \\
& 50 & -0.58 & -0.19 & -0.20 & -0.40 & -0.16 & -0.12 & -4.93 & -4.72 & -4.15 & -3.87 & -3.86 & -3.17 \\
& 100 & -0.03 & -0.96 & -0.61 & -0.73 & -0.78 & -0.54 
& -28.70 & -28.56 & -23.35 & -33.54 & -27.45 & -20.89 \\
& 500 & -3.55 & -2.60 & -1.64 & -2.87 & -2.11 & -1.30 
& -45.39 & -44.31 & -37.38 & -44.87 & -40.73 & -33.85 \\
& 1K & -4.20 & -2.72 & -1.66 & -3.13 & -2.22 & -1.31 
& -45.40 & -44.48 & -39.48 & -44.85 & -41.37 & -35.02 \\
\midrule
\multirow{14}{*}{\rotatebox{90}{AdaFace}}
& \textit{Base} & 85.37 & 94.55 & 96.64 & 92.67 & 96.04 & 97.54 & 85.37 & 94.55 & 96.64 & 92.67 & 96.04 & 97.54 \\
& \textit{Lin} & 85.35 & 94.55 & 96.63 & 92.66 & 96.03 & 97.54 & 85.35 & 94.55 & 96.63 & 92.66 & 96.03 & 97.54 \\
& 1 & -0.10 & -0.03 & +0.00 & -0.04 & +0.01 & -0.02 & -0.15 & -0.05 & -0.05 & -0.20 & -0.08 & -0.06 \\
& 5 & -0.07 & -0.07 & -0.09 & -0.10 & -0.02 & -0.10 & +0.00 & -0.23 & -0.12 & -0.48 & -0.21 & -0.16 \\
& 10 & -0.49 & -0.17 & -0.21 & -0.39 & -0.12 & -0.17 
& -1.19 & -0.71 & -0.37 & -1.17 & -0.54 & -0.38 \\
& 50 & -1.22 & -0.46 & -0.25 & -0.74 & -0.37 & -0.28 
& -0.11 & -2.22 & -1.63 & -1.56 & -1.81 & -1.43 \\
& 100 & -1.45 & -2.02 & -1.21 & -2.52 & -1.49 & -0.94 
& -0.80 & -2.76 & -1.90 & -2.20 & -2.17 & -1.57 \\
& 500 & -8.01 & -8.30 & -5.90 & -8.52 & -6.25 & -4.45 
& -5.99 & -6.27 & -4.44 & -5.87 & -4.63 & -3.21 \\
& 1K & -10.27 & -9.78 & -6.64 & -9.39 & -7.47 & -5.12 
& -7.52 & -8.09 & -5.49 & -7.21 & -5.97 & -4.00 \\
\midrule
\multirow{14}{*}{\rotatebox{90}{MagFace}}
& \textit{Base} & 87.58 & 93.46 & 95.88 & 92.36 & 95.11 & 96.95 & 87.58 & 93.46 & 95.88 & 92.36 & 95.11 & 96.95 \\
& \textit{Lin} & 89.11 & 94.16 & 96.16 & 93.71 & 95.73 & 97.23 & 89.11 & 94.16 & 96.16 & 93.71 & 95.73 & 97.23 \\
& 1 & -0.09 & -0.05 & +0.04 & -0.02 & +0.03 & -0.03 
& +0.03 & -0.12 & +0.01 & -0.07 & -0.09 & -0.02 \\
& 5 & +0.24 & -0.02 & +0.02 & -0.01 & -0.08 & -0.03 & -0.12 & -0.24 & -0.22 & -0.48 & -0.29 & -0.21 \\
& 10 & -0.08 & -0.20 & -0.24 & -0.27 & -0.23 & -0.20 
& -0.29 & -0.62 & -0.54 & -1.13 & -0.55 & -0.50 \\
& 50 & -1.33 & -0.86 & -0.45 & -0.83 & -0.57 & -0.42 
& -3.94 & -2.83 & -2.26 & -3.47 & -2.17 & -1.67 \\
& 100 & -7.51 & -5.66 & -3.86 & -7.41 & -4.71 & -3.29 
& -10.24 & -8.32 & -5.46 & -9.56 & -6.59 & -4.32 \\
& 500 & -26.41 & -23.69 & -16.94 & -22.57 & -17.59 & -12.85 
& -18.98 & -14.84 & -9.42 & -14.73 & -10.42 & -6.57 \\
& 1K & -26.79 & -23.93 & -17.14 & -22.72 & -17.91 & -12.98 
& -19.23 & -14.85 & -9.54 & -15.32 & -10.72 & -6.91 \\
\midrule
\multirow{14}{*}{\rotatebox{90}{QMagFace}}
& \textit{Base} & 89.71 & 94.36 & 96.14 & 93.45 & 95.96 & 97.37 & 89.71 & 94.36 & 96.14 & 93.45 & 95.96 & 97.37 \\
& \textit{Lin} & 89.38 & 94.56 & 96.33 & 93.57 & 96.15 & 97.60 & 89.38 & 94.56 & 96.33 & 93.57 & 96.15 & 97.60 \\
& 1 & -0.07 & +0.01 & -0.06 & -0.08 & -0.02 & -0.03 
& +0.04 & -0.02 & -0.01 & -0.22 & -0.07 & -0.05 \\
& 5 & +0.16 & -0.08 & -0.12 & -0.04 & -0.03 & -0.09 & -0.04 & -0.32 & -0.25 & -0.35 & -0.24 & -0.28 \\
& 10 & +0.00 & -0.23 & -0.15 & -0.07 & -0.23 & -0.19 
& -0.48 & -0.68 & -0.51 & -0.81 & -0.60 & -0.45 \\
& 50 & -1.18 & -0.87 & -0.39 & -0.68 & -0.59 & -0.36 
& -2.93 & -2.70 & -2.04 & -2.92 & -2.19 & -1.68 \\
& 100 & -6.54 & -5.13 & -3.52 & -6.49 & -4.50 & -3.09 
& -8.57 & -7.22 & -4.78 & -7.90 & -5.93 & -4.05 \\
& 500 & -24.76 & -21.42 & -14.90 & -19.76 & -15.40 & -11.00 
& -17.20 & -12.88 & -7.68 & -11.58 & -8.73 & -5.63 \\
& 1K & -25.08 & -21.75 & -15.14 & -20.11 & -15.61 & -11.29 
& -17.23 & -12.69 & -8.14 & -12.25 & -8.97 & -5.86 \\
\bottomrule
\end{tabular}
\end{table*}

\section{Closed-Set Unlearning Evaluation in Open-Set Face Recognition}
As discussed in Section \ref{sec:eval_metrics} and Section \ref{sec:results}, the evaluation metrics adopted in this work are motivated by the open-set nature of FR. Thus, this section provides an additional examination of the limitation of closed-set unlearning metrics and their suitability for evaluating identity-level unlearning in open-set FR.

The face recognition field covers various research directions, including verification, identification, attribute classification (e.g. age and gender) \cite{DBLP:conf/fgr/GundogduUU24, DBLP:conf/pkdd/HossainMLK25, tok2026nearesttargetwrongone, Choi2023TowardsMU}, and face generation \cite{tok2026nearesttargetwrongone}. Existing machine unlearning approaches for FR often formulate the problem from a classification perspective \cite{DBLP:conf/cikm/LeeW23, 10426386Compromising, DBLP:conf/wacv/Zakharov26, DBLP:conf/icb/ShivamNRG25}, where conventional closed-set metrics such as forget and retain accuracy are used to assess the unlearning performance. However, in open-set FR, verification relies on identity-discriminative embeddings rather than predictions over a fixed set of training identities. Therefore, changes in classification accuracy do not necessarily reflect the extent to which identity-related information has been altered in the embedding space. This can be one of the main reasons for the use of embedding-level evaluation metrics, such as the FS, which statistically measures changes in the probability distributions of cosine similarities for unlearned identities before and after unlearning. Unlike classification-based metrics, FS directly evaluates changes at the representation level, making it more suitable for unlearning evaluation in open-set settings.  

Membership Inference Attack (MIA) is another commonly used metric for evaluating machine unlearning. Originally developed to assess training-set membership leakage, MIA has been utilized to determine whether unlearned samples maintain detectable membership signals after unlearning. Ideally, their behavior should resemble that of samples excluded from training. Several unlearning studies have used MIA \cite{DBLP:conf/aaai/FosterSB24, DBLP:conf/iclr/FosterSB24, DBLP:conf/icb/ShivamNRG25}, mostly relying on classification-based signals, such as prediction confidence, entropy or loss, to distinguish members from non-members. Alternatively, De Min \etal, in their paper titled \textit{"Unlearning Personal Data from a Single Image,"} published in Transactions on Machine Learning Research (TMLR, 2025), used a reference-based MIA (RMIA) to assess identity-level unlearning but in face-related tasks. However, these approaches require either classification-dependent signals or additional reference models, which are not directly aligned with embedding-based (open-set) evaluation considered in this work. Moreover, RMIA requires reference models trained without the forget data, which introduces substantial additional computational overhead for large-scale FR models. 

We therefore do not consider MIA in this work. Evaluating it faithfully in our setting would require RMIA, whose reference models alone entail several months of additional training at the scale considered. Moreover, RMIA is formulated for classification outputs and would itself have to be adapted to operate on face embeddings. Adapting the attack to identity-level unlearning in open-set FR is likewise out of scope here, and we leave such an evaluation to future work.

\end{document}